\documentclass[lettersize,journal]{IEEEtran}
\usepackage{amsmath,amsfonts}
\usepackage{algorithmic}
\usepackage{algorithm}
\usepackage{array}
\usepackage[caption=false,font=normalsize,labelfont=sf,textfont=sf]{subfig}
\usepackage{textcomp}
\usepackage{stfloats}
\usepackage{url}
\usepackage{verbatim}
\usepackage{graphicx}
\usepackage{cite}
\usepackage{xspace}
\makeatletter
\DeclareRobustCommand\onedot{\futurelet\@let@token\@onedot}
\def\@onedot{\ifx\@let@token.\else.\null\fi\xspace}
\def\eg{\emph{e.g}\onedot} 
\def\ie{\emph{i.e}\onedot} 
\def\cf{\emph{c.f}\onedot}

\makeatother
\usepackage{threeparttable}
\usepackage{color}
\usepackage{colortbl}
\definecolor{mygray1}{gray}{.9}
\definecolor{mygray2}{gray}{.8}
\definecolor{mygray3}{gray}{.6}
\usepackage{makecell}
\usepackage{multirow}
\usepackage{multicol}
\usepackage{booktabs}
\usepackage{amsmath}
\usepackage{amssymb}
\usepackage{pifont}
\usepackage{booktabs}
\usepackage{bbding}
\usepackage{array}
\usepackage{amsmath,bm}
\newcommand{\PreserveBackslash}[1]{\let\temp=\\#1\let\\=\temp}
\newcolumntype{C}[1]{>{\PreserveBackslash\centering}p{#1}}
\newcolumntype{R}[1]{>{\PreserveBackslash\raggedleft}p{#1}}
\newcolumntype{L}[1]{>{\PreserveBackslash\raggedright}p{#1}}

\usepackage{algorithm}
\usepackage{algorithmic}

\usepackage{newfloat}
\usepackage{listings}

\usepackage{amsfonts}
\begin{document}

\title{Omni-frequency Channel-selection Representations \\for Unsupervised Anomaly Detection}

\author{Yufei Liang$^*$, Jiangning Zhang$^*$, Shiwei Zhao, Runze Wu, Yong Liu, and Shuwen Pan
        % <-this % stops a space
\thanks{Yufei Liang and Jiangning Zhang contributed equally. Yong Liu is the corresponding author.}

\thanks{Yufei Liang, Jiangning Zhang, and Yong Liu are with the Laboratory of Advanced Perception on Robotics and Intelligent Learning, College of Control Science and Enginneering, Zhejiang University, Hangzhou 310027, China; Email: 22032139@zju.edu.cn, 186368@zju.edu.cn, yongliu@iipc.zju.edu.cn.}

\thanks{Shiwei Zhao and Runze Wu are with the Fuxi AI Lab, NetEase Games, Hangzhou 310012, China; Email: zhaoshiwei@corp.netease.com, wurunze1@corp.netease.com.}

\thanks{Shuwen Pan is with the Discipline of Control Science and Engineering, School of Information and Electrical Engineering, Zhejiang University City College, Hangzhou 310015, China; Email: pansw@zucc.edu.cn}

}

% The paper headers
% \markboth{Journal of \LaTeX\ Class Files,~Vol.~14, No.~8, August~2021}%
% {Shell \MakeLowercase{\textit{et al.}}: A Sample Article Using IEEEtran.cls for IEEE Journals}

%\IEEEpubid{0000--0000/00\$00.00~\copyright~2021 IEEE}
% Remember, if you use this you must call \IEEEpubidadjcol in the second
% column for its text to clear the IEEEpubid mark.

\maketitle

\begin{abstract}
Density-based and classification-based methods have ruled unsupervised anomaly detection in recent years, while reconstruction-based methods are rarely mentioned for the poor reconstruction ability and low performance. However, the latter requires \emph{no costly extra training samples for the unsupervised training} that is more practical, so this paper focuses on improving reconstruction-based method and proposes a novel \textit{\textbf{O}}mni-frequency \textit{\textbf{C}}hannel-selection \textit{\textbf{R}}econstruction (OCR-GAN) network to handle sensory anomaly detection task in a perspective of frequency. Concretely, we propose a Frequency Decoupling (FD) module to decouple the input image into different frequency components and model the reconstruction process as a combination of parallel omni-frequency image restorations, as we observe a significant difference in the frequency distribution of normal and abnormal images. Given the correlation among multiple frequencies, we further propose a Channel Selection (CS) module that performs frequency interaction among different encoders by adaptively selecting different channels. Abundant experiments demonstrate the effectiveness and superiority of our approach over different kinds of methods, \eg, achieving a new state-of-the-art \textbf{98.3} detection AUC on the MVTec AD dataset without extra training data that markedly surpasses the reconstruction-based baseline by \textbf{+38.1}$\uparrow$ and the current SOTA by \textbf{+0.3}$\uparrow$. 
% Source code is available at \url{https://github.com/zhangzjn/OCR-GAN}. 
% The source code is available in the additional materials. 
\end{abstract}


\section{Introduction}\label{sec:introduction}
\begin{figure}[t]
    \centering
    \includegraphics[width=1.0\columnwidth]{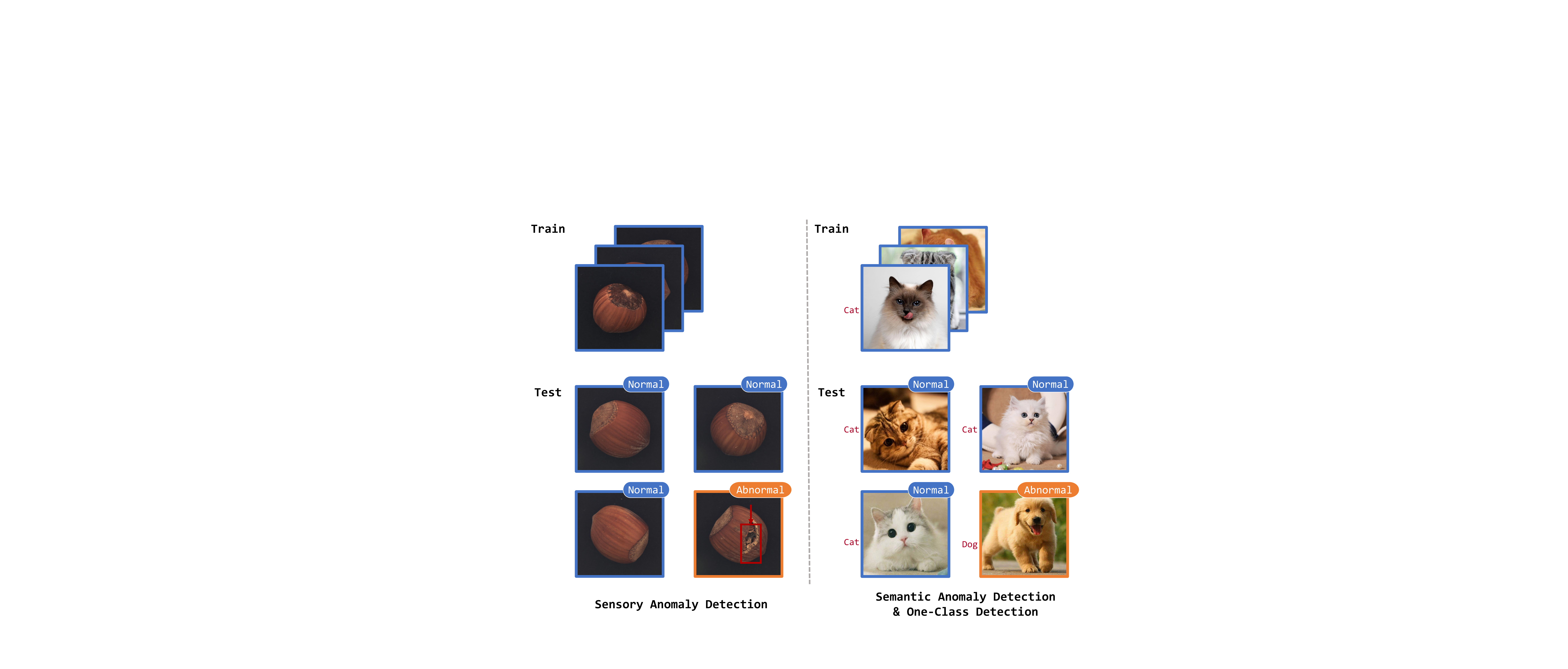}
    \caption{Illustrations of sensory anomaly detection (\textbf{Left}) and semantic anomaly detection (\textbf{Right}) .}
    \label{fig:AD}
\end{figure}

\IEEEPARstart{A}{nomaly} detection is a binary classification task to distinguish whether a given image deviates from the predefined normality, which is an essential task in visual image understanding that has various applications in the real world, \eg, novelty detection~\cite{nilsback2008automated}, product quality monitoring based on industrial images~\cite{bergmann2019mvtec}, automatic defect restoration~\cite{wang2012archive}, human health monitoring~\cite{li2018thoracic} and video surveillance~\cite{liu2018classifier,sultani2018real,lv2021localizing,jardim2019domain}. In real-world applications, anomaly detection tasks can be divided into sensory AD (Fig.~\ref{fig:AD}a) and semantic AD (Fig.~\ref{fig:AD}b): the former only suffers from covariate shift~\cite{yang2021generalized} without semantic shift, while the later is the opposite. Most anomalies appear in the form of defects in the sensory AD, such as the normal defect detection task in MVTec AD~\cite{bergmann2019mvtec} and KolektorSDD~\cite{Tabernik2019JIM} datasets. However, the semantic AD task detects images with label shifts, assuming that normal and abnormal come from different semantic distributions, such as the one-class detection task in CIFAR-10~\cite{krizhevsky2009learning}. This work focuses on solving the sensory AD task but also evaluates on the related semantic AD dataset.

\begin{figure*}[t]
    \centering
    \includegraphics[width=1.0\linewidth]{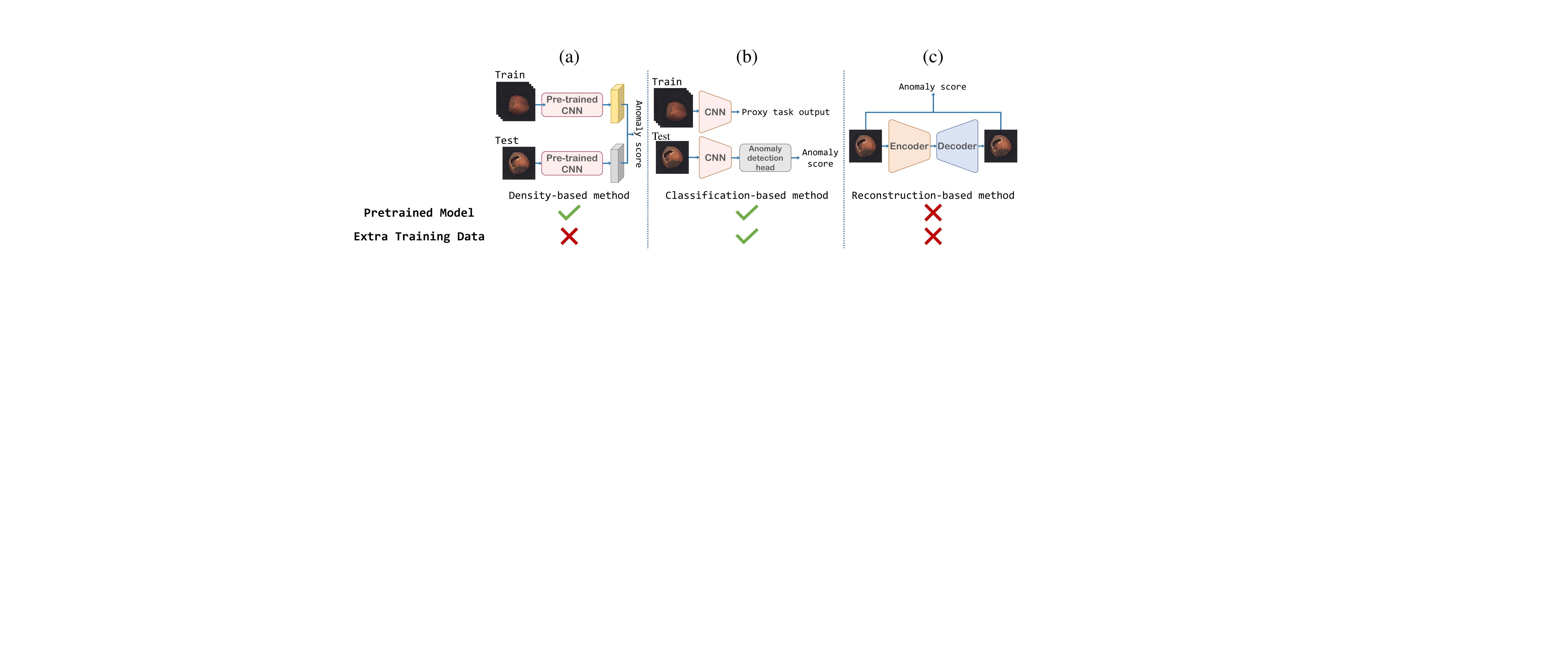}
    \caption{Pipeline illustrations of three kinds of unsupervised anomaly detection methods in column. Bottom two rows indicate whether \emph{Pretrained Model} and \emph{Extra Training Data} are used for each kind of method.}
    \label{fig:un_methods}
\end{figure*}

In anomaly detection, obtaining abnormal samples and detecting novel abnormalities are time-consuming and costly objects that force us to develop unsupervised methods for more practical applications. Current unsupervised anomaly detection methods are mainly divided into three categories: density-based (Fig.~\ref{fig:un_methods}a), classification-based (Fig.~\ref{fig:un_methods}b) and reconstruction-based (Fig.~\ref{fig:un_methods}c) methods. \emph{\textbf{a)} Density-based methods} generally employ a pre-trained model to extract meaningful vectors of the input image. The anomaly score can be obtained by calculating the similarity between the embedding representation of the test image and the reference density distribution. This kind of method~\cite{cohen2020sub,yi2020patch,defard2021padim} achieves a high AUC score on the popular MVTec AD~\cite{bergmann2019mvtec} dataset, but they \emph{need pre-trained models and are insufficient for the model interpretability}. \emph{\textbf{b)} Classification-based methods} try to find the classification boundaries of normal data. Self-supervised methods are representative of classification-based methods, which use the model trained by the proxy task to detect anomalies. Thus, self-supervised methods \emph{rely on how well the proxy tasks match the test data}. For example, CutPaste~\cite{li2021cutpaste} performs well in anomaly detection on the MVTec AD dataset. However, it is difficult for this method to perform well on other datasets. Also, these methods \emph{rely on pre-trained models and extra training data}. \emph{\textbf{c)} Reconstruction-based methods}~\cite{ravanbakhsh2017abnormal,nguyen2019anomaly,xia2021discriminative, hu2020lightweight} contain a generator to reconstruct the input image, and the anomaly score is the more interpretable reconstruction error. \emph{These methods do not need pre-trained models and extra training data}. However, current reconstruction-based methods without extra training data are much less expressive than other methods. In summary, current unsupervised anomaly detection approaches are still suffering from two main challenges: \emph{\textbf{(1))}} Some works achieve high AUC score but require abnormal samples or extra training data that are hard to obtain and costly for practical use. \emph{\textbf{(2))}} Current reconstruction-based methods are more practical and do not need pre-trained models and extra training data but suffer from low performance. Although our approach borrows from self-supervised methods for constructing pseudo-anomaly data, this paper focuses on improving the reconstruction-based method as it requires no extra training data and only normal samples which is more practical.

\begin{figure*}[t]
    \centering
    \includegraphics[width=1.0\linewidth]{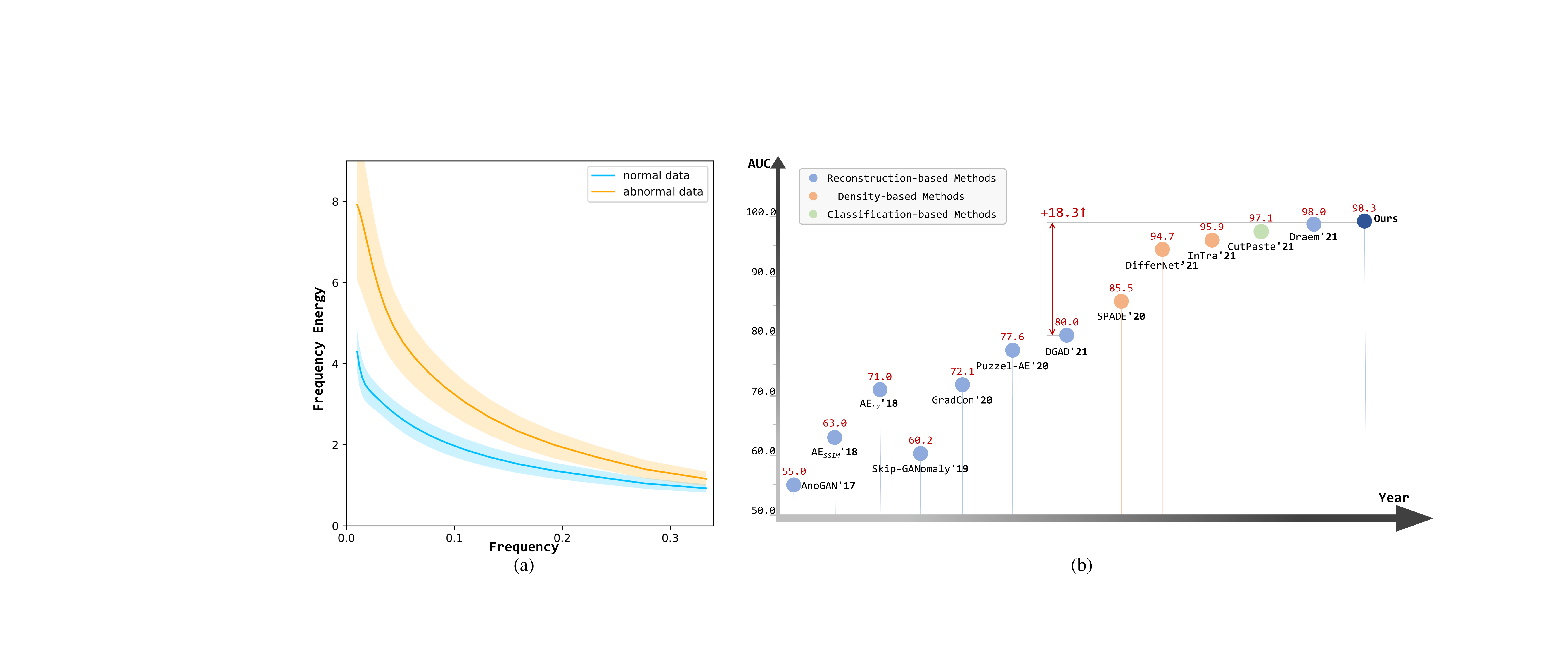}
    \caption{(a) \textbf{Energy distribution} with frequencies for normal and abnormal samples in MVTec AD dataset, and the shadow represents standard deviation. \emph{Normal and abnormal data have noticeable frequency distribution differences.} (b) \textbf{Development of three kinds of methods}. Our approach surpasses the SOTA reconstruction-based method without extra training data by a large margin, i.e., +18.3$\uparrow$. Note that the current SOTA Draem~\cite{zavrtanik2021draem} is not a classical reconstruction-based method, which requires a new training strategy and extra training data.}
    \label{fig:motivation}
\end{figure*}
To improve the performance of the reconstruction-based method, we need to enhance the reconstruction ability of the generator for the anomaly detection task. For an image, different frequency bands contain different types of information, \eg, low frequency represents more semantic information while high frequency represents more detailed texture information. Also, we find that the model performance can be improved from the frequency domain perspective in many computer vision tasks, \eg, in image super-resolution task, \cite{li2020learning} separates the different frequency components to compensate for the loss of information in different frequency bands of real LR images to improve the performance of the model. Motivated by the idea, we analyze the frequency distribution of normal and abnormal images in the anomaly detection task. As shown in Fig.~\ref{fig:motivation}(a), we count the frequency energy distribution of normal and abnormal images, as the energy distribution of the Fourier-transformed image is reflected in the amplitude spectrum. We re-analyze this paradigm and find that normal and abnormal samples have different frequency distributions in sensory AD. So it may be difficult and unsuitable for only one generator to learn the full-frequency reconstruction of the RGB image. Therefore, we propose an anomaly detection framework using multiple frequency branches to reconstruct information from different frequency bands respectively. In order to differentiate the use of information from different frequency bands, we propose an effective \emph{Frequency Decoupling} (FD) module to pre-obtain omni-frequency representation of the input image and use parallel generators to reconstruct images of multiple frequencies. Considering the model efficiency, we conduct experiments with 2 or 3 frequency branches in this paper. Different frequency branches in the framework are independent by default. However, an image contains information in multiple frequency bands, and the information in different frequency bands is not completely unrelated to each other but complementary in the real world. So, we design a tailored \emph{Channel Selection} (CS) module to further realize omni-frequency interaction among multiple branches that can adaptively select different channel features. Based on the above modules and the baseline Skip-GANomaly~\cite{akccay2019skip}, we propose a novel \textit{\textbf{O}}mni-frequency \textit{\textbf{C}}hannel-selection \textit{\textbf{R}}econstruction (OCR-GAN) network. Our method achieves state-of-the-art (SOTA) results on multiple public datasets consistently. Specifically, our OCR-GAN improves +0.3$\uparrow$ than current SOTA method Draem~\cite{zavrtanik2021draem} and significantly +18.3$\uparrow$ than SOTA reconstruction-based DGAD~\cite{xia2021discriminative} without extra training data on MVTec AD in Fig.~\ref{fig:motivation}(b), \emph{which strongly proves that the reconstruction-based method can also perform well even without extra training data and pre-trained models}. To the best of our knowledge, this paper is the first attempt to explore omni-frequency information with reconstruction-based anomaly detection method. Our main contributions can be summarized as follows:
\begin{itemize}
    \item We rethink the difference between normal and abnormal images from the frequency domain perspective and propose a novel framework for anomaly detection based on omni-frequency reconstruction.
    \item We propose an effective FD module to obtain different frequency band information of the image that enables the omni-frequency reconstruction by multiple branches.
    \item We propose a CS module to realize omni-frequency interaction among multiple branches and adaptive selection of different channel features.
    \item Abundant experiments demonstrate the superiority of our OCR-GAN over SOTA methods, \eg, we achieve a new SOTA \textbf{98.3} detection AUC on the MVTec AD dataset without extra training data, which markedly surpasses the SOTA reconstruction-based method without extra training data by \textbf{+18.3}$\uparrow$ and the SOTA method by \textbf{+0.3}$\uparrow$.
\end{itemize}

The remainder of the paper is organized as follows. In Sec.~\ref{sec:related_work}, we review some related works. Details of the proposed OCR-GAN method are given in Sec.~\ref{sec:our_approach}. Experimental results are presented in Sec.~\ref{sec:experiments}. And we conclude the paper with discussion and summary in Sec.~\ref{sec:conclusion}.

\section{Related Work}\label{sec:related_work}
% Most methods of sensory AD and semantic AD are shared, but methods of sensory AD focus more on the local information of the image and the internal information of the neural network. 
Anomaly detection methods can be mainly divided into density-based, classification-based and reconstruction-based methods as follows.
\subsection{Density-based methods}
Density-based methods build a density estimation model for the distribution of normal training data. And this kind of method assumes that normal data have a higher likelihood under this model than abnormal data during inference. Parameter density estimation assumes that the density of normal data can be represented by some reference distribution.
%\subsection{Embedding-based methods}
A pre-trained network is used to extract meaningful vectors representing the whole image or patch image for anomaly detection. The similarity between the representation vector of the test image and the reference vector is set as anomaly score. Some researches~\cite{andrews2016transfer, nazare2018pre, bergman2019classification, rippel2021modeling, yang2020dfr, zeng2021reference} train the model on the entire image, while works~\cite{zhang2013region, yi2020patch, defard2021padim, bergmann2020uninformed} on the patch image.
The normal distribution reference can be the parameter of the Gaussian distribution of the normal image embedding vectors~\cite{chen2021unsupervised, defard2021padim}, the mixed Gaussian distribution~\cite{eskin2000anomaly,redner1984mixture}, the Poisson distribution~\cite{turcotte2016poisson}, the center of the sphere containing the embedding from normal images~\cite{ruff2018deep, yi2020patch}, the entire set of normal embedding vectors~\cite{cohen2020sub, bergman2020deep}, the feature of the last layer in the network~\cite{chen2001one, sabokrou2018deep}, or the mid-level feature representation~\cite{roth2022towards}. Mahalanobis distance is used to calculate the anomaly score between the embedding vector of the test image and the reference vector of the normal training distribution. 
The PaDIM\cite{defard2021padim} elaborate that the density-based model (\ie, embedding similarity-based model) lacks interpretability. This method only performs anomaly detection and gives promising results. However, it lacks interpretability as it is impossible to know which part of an anomalous image is responsible for a high anomaly score. PaDIM interprets the location of anomalies by detecting them in the patch, but this introduces a large amount of computation.
%These methods have achieved good performance recently, but they lack interpretability that it is difficult to clearly distinguish which part of the image causes high abnormal scores. 
Also, this kind of method requires the pre-trained model for extracting vectors that is less practical for various real scenarios.

Another method of density estimation is normalizing flows. Normalizing flows are used to learn bijective transformations between data distributions with a special property. DifferNet~\cite{rudolph2021same} using normalizing flows to estimate the precise likelihood . Since flow-based methods have no dimensional reduction, the computation cost is significant. And this kind of method also needs pre-trained models to extract features.

\begin{figure*}[ht]
    \centering
    \includegraphics[width=1\linewidth]{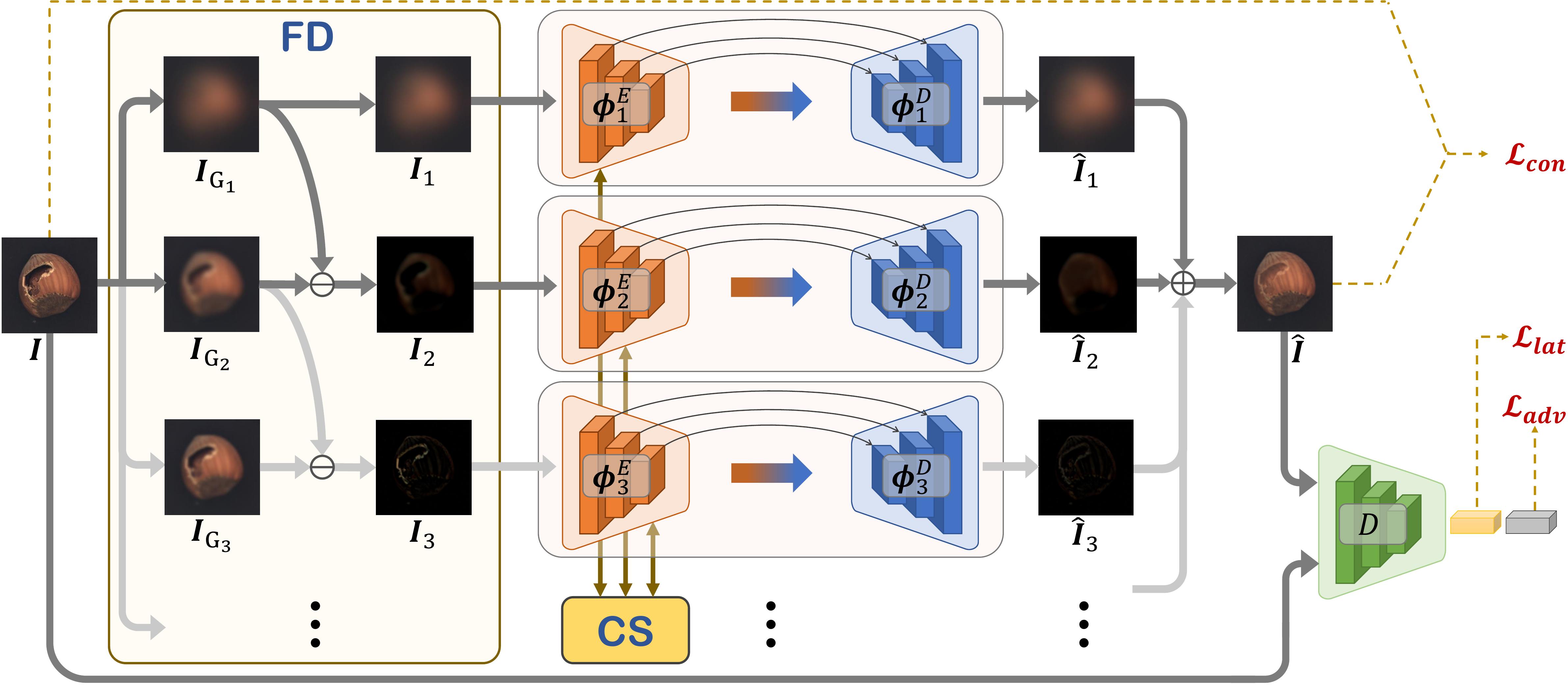}
    \caption{\textbf{Overview of proposed OCR-GAN}. Input image $\bm{I}$ goes through Frequency Decoupling (FD) module to obtain omni-frequency images $\{\bm{I}_1, \bm{I}_2, \dots\}$ from pre-processed Gaussian images $\{\bm{I}_{G_1}, \bm{I}_{G_2}, \dots\}$. Then $\{\bm{I}_1, \bm{I}_2, \dots\}$ are fed into multiple generators $\{\bm{\phi}_1, \bm{\phi}_2, \dots\}$ to reconstruct corresponding images $\{\hat{\bm{I}}_1, \hat{\bm{I}}_2, \dots\}$, which are added to obtain the final output $\hat{\bm{I}}$. The proposed Channel Selection (CS) module performs omni-frequency interaction among different encoders, \ie, $\{\bm{\phi}_1^E, \bm{\phi}_2^E, \dots\}$.}
    \label{fig:network}
\end{figure*}

\subsection{Classification-based methods}
Classification-based methods~\cite{ju2015image} try to find the classification boundaries of normal data. DeepSVDD~\cite{ruff2018deep} first introduces one-class classification to anomaly detection. 
Moreover, there are some self-supervised learning methods to design good proxy tasks to help the model detect anomalies from normal samples. One classical self-supervised anomaly detection method is isolation forest~\cite{liu2008isolation}. 
Other proxy tasks for self-supervised anomaly detection methods include image transformation prediction~\cite{golan2018deep,bergman2019classification }, contrastive learning~\cite{tack2020csi} and proxy binary classification~\cite{li2021cutpaste}. \cite{li2021cutpaste} uses data augmentation to generate pseudo-anomaly data and then does a binary classification proxy task with normal training samples to train the feature extraction model. The self-supervised method relies on the design of proxy tasks, which is difficult to perform well on multiple data sets. Representation learning performance in self-supervised methods relies on the ability of network feature extraction. Therefore, such methods usually use a pre-trained model as the feature extraction network.

\subsection{Reconstruction-based methods}
One of the reconstruction-based methods is sparse reconstruction which assumes that normal samples can be reconstructed with a limited number of basis functions while abnormal samples are more expensive to reconstruct. $L_1$ norm-based kernel PCA~\cite{xiao2013l1} and low-rank embedded networks~\cite{jiang2021lren} are belong to sparse reconstruction methods.

 The reconstruction method is intuitive and easy to understand. Abnormal images would get higher reconstruction errors as they have a different data distribution than normal images. The autoencoder (AE)~\cite{kingma2014auto} and generative adversarial networks (GAN)~\cite{goodfellow2014generative} can reconstruct samples from the normal training data. ~\cite{bergmann2018improving} propose to use an autoencoder for the reconstruction process and structural similarity to measure reconstruction error. Some studies~\cite{goodfellow2014generative, pathak2016context} have shown that using adversarial network training would improve generation results. Moreover, GAN-based methods have more suitable metrics that can play the role of anomaly score, \eg, output of the discriminator~\cite{sabokrou2018adversarially, akcay2018ganomaly} and latent space distance~\cite{abati2019latent, akcay2018ganomaly, akccay2019skip}.  %It is difficult to ensure the poor reconstruction for abnormal samples as the capacity of the generator is strong, so 
 For more accurate anomaly detection, OCGAN~\cite{perera2019ocgan} uses a denoising autoencoder, latent discriminator, visual discriminator, and classifier to ensure that any example generated from the learned latent space is indeed from the normal class. For GAN-based methods, the discriminator is usually used to distinguish the reconstructed image from the original image, but OGNet~\cite{zaheer2020old} redefines the role of the discriminator that is used to distinguish reconstructed images of different qualities. Recently, ~\cite{kwon2020backpropagated} utilize backpropagated gradients as representations to characterize anomalies, and %DGAD~\cite{xia2021discriminative} learns representation by the guidance of the discriminator to improve the model performance. 
 The generation ability of the generator has a significant influence on the effect of the reconstruction-based method, so ~\cite{han2021gan} propose to construct GAN ensembles for anomaly detection as GAN ensembles often outperform the single GAN. %And ~\cite{liu2019multistage} propose multistage GAN to detect fabric defect. 
 It is challenging to ensure poor reconstruction for abnormal samples as the capacity of the generator is strong. Thus these methods perform poorly in sensory detection. These methods indiscriminately reconstruct all frequencies of the RGB image that may be difficult for the generator, leading to poor results in anomaly detection. Also, we find that normal and abnormal samples have different frequency distributions, so we propose a new paradigm that uses parallel branches to reconstruct omni-frequency images.

\section{Our Approach}\label{sec:our_approach}
\subsection{Overview}
In this section, we aim at improving the current reconstruction-based approach without extra training data and designing a generalized network for anomaly detection. As the difference between normal and abnormal images varies in different frequency bands, we perform anomaly detection from the perspective of the frequency domain. As shown in Fig.~\ref{fig:network}, our method derives from a frequency-decoupling idea that comprises multiple generators, \ie, $G$=$\{\bm{\phi}_1, \bm{\phi}_2, \dots\}$, to reconstruct omni-frequency images $\{\hat{\bm{I}}_1, \hat{\bm{I}}_2, \dots\}$, which is trained alternately with a discriminator $D$ to further boost the model performance. Concretely, we propose an effective FD module to decouple the input image $\bm{I}$ to omni-frequency images $\{\bm{I}_1, \bm{I}_2, \dots\}$ and a CS module to realize omni-frequency interaction by adaptively selecting channels among encoders $\{\bm{\phi}_1^E, \bm{\phi}_2^E, \dots\}$. When the model finishes the training, the abnormal images would be poorly reconstructed and get higher anomaly scores than normal images. 
\subsection{Frequency Decoupling}
Pixel distributions reflect the spatial frequency of the image. Different frequency components contain different information, \eg, the low frequency of the image contains more semantic information while the high frequency includes more details and texture information. As previously mentioned, normal and abnormal images have obvious frequency distribution differences, which derive from the abnormal elements in abnormal data, \eg, holes, cracks, and scratches in the MVTec AD dataset. For a more thorough analysis, we counted the frequency energy distribution of normal and abnormal images. The energy distribution of the Fourier-transformed image is reflected in the amplitude spectrum. And Fig.~\ref{fig:motivation}(a) also shows the difference between normal and abnormal images in the frequency domain. Therefore, we consider that the importance of information in different frequency bands varies in anomaly detection tasks, especially sensory anomaly detection.

Motivated by the difference in the frequency distribution of normal and abnormal images shown in Fig.~\ref{fig:motivation}(a), we propose a tailored Frequency Decoupling (FD) module to pre-obtain informative omni-frequency representations. Specifically, FD contains the following three processes. 

\noindent\textbf{(1)} Convolving original image $\bm{I}$ with the Gaussian kernel $\bm{Gau}_1$:

\begin{equation}
    \bm{Gau}_1 = \frac{1}{256}\left[\begin{array}{ccccc}
    1&4&6&4&1\\
    4&16&24&16&4\\
    6&24&36&24&6\\
    4&16&24&16&4\\
    1&4&6&4&1\\
    \end{array}\right],
\end{equation}
and then removing even rows and columns of the blurred image to obtain intermediate down-sampled image $\bm{I}_{blur}$.
\begin{figure*}[t]
    \centering
    \includegraphics[width=0.8\linewidth]{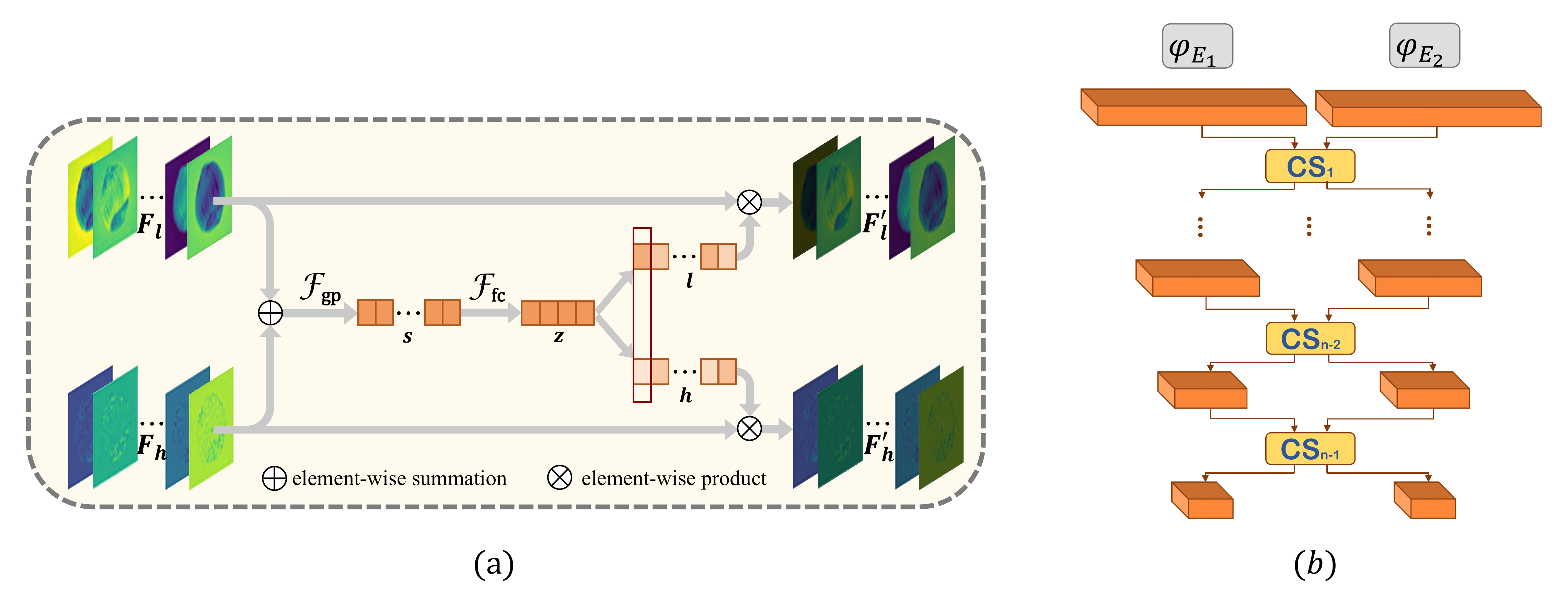}
    \caption{\textbf{(a) Schematic diagram of CS}. For simplicity, a two-branch case is shown here that $F_l$ and $F_h$ represent low-/high-frequency features, \ie, features in branch one and two. Augmented $F_{l}^{'}$ and $F_{h}^{'}$ are fed into following layers. \textbf{(b) Integration of CS to the framework}. The CS module is used at each stage of encoding.}
    \label{fig:atten}
\end{figure*}

\noindent\textbf{(2)} $\bm{I}_{blur}$ would be exactly one-quarter the area of $\bm{I}$ and goes through a $\times$2$\uparrow$ up-sampling to restore the original resolution, with the new even rows and columns filled with zeros. Then a similar convolution operation with the Gaussian kernel $\bm{Gau}_2 = 4 * \bm{Gau}_1$ is applied to approximate missing pixels, \ie, zeros in even rows and columns, and we obtain first-level blurred image $\bm{I}_{G_{N-1}}$, where $n$ represents the branch number (The smaller the $N$ is, the less high-frequency information) and $\bm{I}_{G_{N}}$ is initialized with $\bm{I}$, denoted as:

\begin{equation}
\begin{aligned}
    \bm{I}_{G_{N}}      &= \bm{I},\\
    \bm{I}_{G_{N-1}}    &= \text{Up}(\text{Down}(\bm{I}_{G_{N}} * \bm{Gau}_1)) * \bm{Gau}_2,
\end{aligned}
\end{equation}
where $*$ means the convolution operation, while $\text{Down}$ and $\text{Up}$ are above-mentioned down-sampling and up-sampling operations. We would obtain a set of blurred images $\{\bm{I}_{G_{1}}, \bm{I}_{G_{2}}, \dots, \bm{I}_{G_{n}}\}$ by repeating the above processes. 

\noindent\textbf{(3)} The blurred images $\bm{I}_{G_{n}}, n=1, 2, \dots, N-1$ lost some high-frequency information in varying degrees, and we further calculate the difference between adjacent images to obtain omni-frequency images:

\begin{equation}
\begin{aligned}
    \bm{I}_1 &= \bm{I}_{G_{1}},\\
    \bm{I}_2 &= \bm{I}_{G_{1}} - \bm{I}_{G_{2}},\\
    & ~~\vdots \\
    \bm{I}_N &= \bm{I}_{G_{N-1}} - \bm{I}_{G_{N}}.
\end{aligned}
\end{equation}
FD can be effectively applied to frequency-sensitive tasks such as anomaly detection by decoupling different-frequency components as needed, and we set branch number two in the paper by default. Fig.~\ref{fig:network} shows that different frequency components are reconstructed by multiple independent generators.

\subsection{Channel Selection}
Different frequency branches are relatively independent in our anomaly detection framework with only the FD module, which goes against the objective fact that different frequencies complement each other. Besides, as shown in Fig.~\ref{fig:atten} (a), the features of different channels in omni-frequency features are various. Attention~\cite{li2019selective} can help us to achieve the selection of information in different frequency bands. Therefore, we design a novel Channel Selection (CS) module to realize omni-frequency interaction among multiple branches and adaptive selection of different channel features. Fig.~\ref{fig:atten} shows the detailed structure of the CS module with a two-branch case that contains low and high-frequency features, but it is easy to extend to multiple branches. Concretely, for the given two feature maps $\bm{F}_h \in \mathbb{R}^{H\times W\times C}$ with high-frequency information and $\bm{F}_l \in \mathbb{R}^{H\times W\times C}$ with low-frequency information, we fuse them via an element-wise summation:
\begin{equation}
    \bm{F} = \bm{F}_l + \bm{F}_h.
\end{equation}
Then we apply Global Average Pooling to embed the global information and obtain channel-wise statistics $\bm{z}^1 \in {R}^{C}$:
\begin{equation}
    \bm{z}^1_c = \mathcal{F}_{GAP}(\bm{F}_c) = \frac{1}{H \times W}\sum^H_{i=1}\sum^W_{j=1}\bm{F}_c(i, j),
\end{equation}
where $c$ is the $c$-th channel with $C$ channels totally. After that, we use a fully connected layer to reduce the dimension of the embedded $\bm{z}^1$ from $C$ to $d$ and obtain $\bm{z}^2 \in {R}^{d}$, which is able to provide a precise and adaptive selection:
\begin{equation}
    \bm{z}^2 = \mathcal{F}_{FC}(\bm{z}^1).
\end{equation}
Finally, we use compact feature descriptor $\bm{z}^2$ to regress $c$-th channel attentions for different frequency branch by:
\begin{equation}
\begin{aligned}
    \bm{l}_c &= \frac{e^{\bm{L}_{c} \bm{z}^2}}{e^{\bm{L}_c \bm{z}^2}+e^{\bm{H}_c \bm{z}^2}},\\
    \bm{h}_c &= \frac{e^{\bm{H}_{c} \bm{z}^2}}{e^{\bm{L}_c \bm{z}^2}+e^{\bm{H}_c \bm{z}^2}},
\end{aligned}
\label{eqa:1}
\end{equation}
where $\bm{L}, \bm{H} \in {R}^{C\times d}$ represent the parameter weights, while $\bm{l}$ and $\bm{h}$ denote the channel attention vectors for $\bm{F_l}$ and $\bm{F_h}$. The augmented feature maps $\bm{F}_l^{'}$ and $\bm{F}_h^{'}$ are obtained through the channel attention operations as follows:
\begin{equation}
\begin{aligned}
    \bm{F}^{'}_{l_c} &= \bm{l}_c \cdot \bm{F}_{l_c},\\
    \bm{F}^{'}_{h_c} &= \bm{h}_c \cdot \bm{F}_{h_c}.
\end{aligned}
\end{equation}
CS module augments feature maps of different frequency branches by adaptively selecting channels, \ie, using $\bm{l}$ and $\bm{h}$ to re-weight $\bm{F}_l^{'}$ and $\bm{F}_h^{'}$. Note that the attention vectors of two branches complement each other, \ie, $\bm{l}_c$ + $\bm{h}_c$ = 1, and this module can be easily extended to multiple branches by adding corresponding weights in Equation~\ref{eqa:1}. 

The CS module is applied to the encoding stage of the generator. We use different generators to encode different frequency band information. As shown in Fig.~\ref{fig:channel} (b), the feature maps of each layer of the individual encoders are used as input to the CS module, and the output of the CS module is used as input to the next layer of the individual encoders. In general, the CS module is used at each stage of encoding.

\subsection{Training}
\begin{figure}[t]
    \centering
    \includegraphics[width=1\columnwidth]{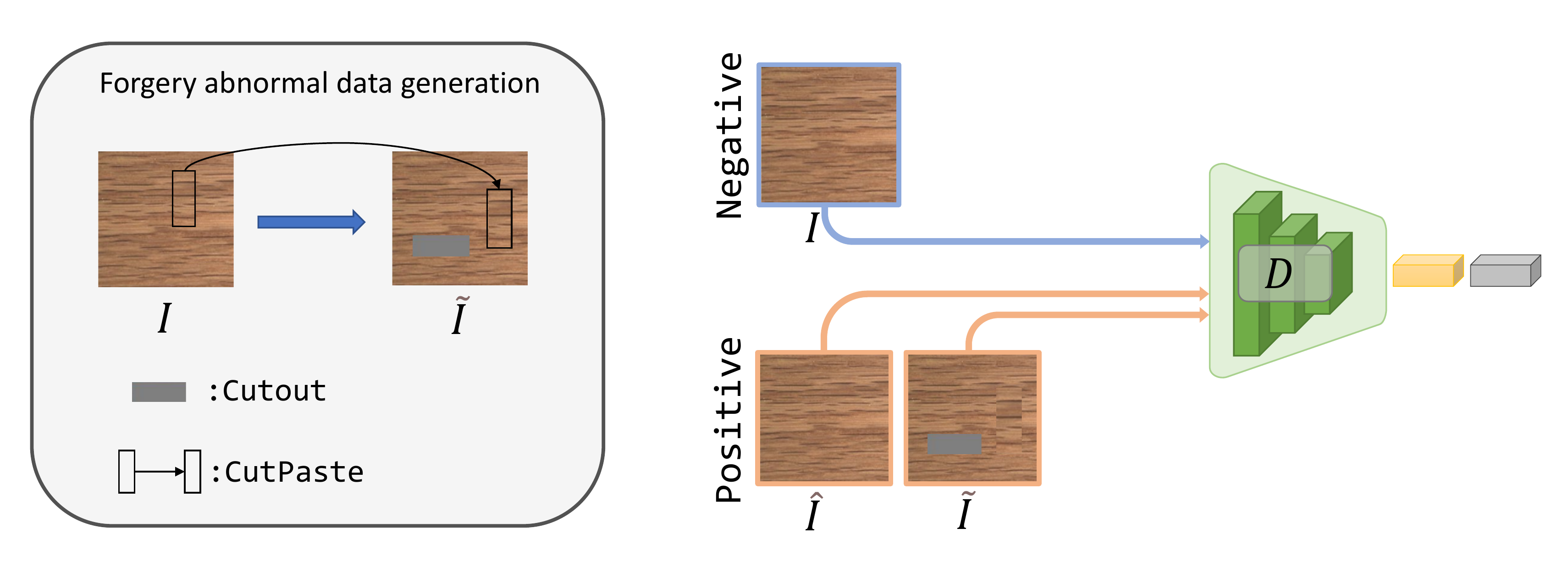}
    \caption{Schematic diagram of using abnormal forgery samples to assist the training process.}
    \label{fig:forgery}
\end{figure}

Our OCR-GAN is trained from scratch end-to-end with only normal samples and tested with both normal and abnormal samples. We expect the GAN to correctly reconstruct normal samples both in the image and latent vector space. Consistent with the previous reconstruction-based methods, our OCR-GAN assumes that out-of-distribution (\ie, anomaly pixels) cannot be well reconstructed as the model is never trained on abnormal samples. Therefore, the difference between the reconstructed image and the input image in either image space or latent space is much greater for abnormal samples. Moreover, inspired by CutPaste~\cite{li2021cutpaste}, we also use data augmentation to generate forgery abnormal samples to assist the training process. Specifically, for normal images used in the training stage, we apply both CutPaste and CutOut~\cite{cutout} on each normal image to generate the forgery abnormal data. As shown in Fig.~\ref{fig:forgery}, the forgery abnormal samples generated by data augmentation and the samples generated by the generator are both used as positive inputs to the discriminator $D$, while original normal samples are used as negative inputs. As shown in Fig.~\ref{fig:network}, we adopt three losses for model training to ensure that OCR-GAN can well reconstruct normal samples.

\noindent{\textbf{Content Loss.}} The first term $\mathcal{L}_{con}$ ensures accurate reconstruction between the input normal image $\bm{I}$ and the reconstructed image $\hat{\bm{I}}$ by $\ell_1$ error:
\begin{equation}
\begin{aligned}
    \mathcal{L}_{con} &= \mathbb{E}_{\bm{I}\sim p_n}[\bm{I}-\hat{\bm{I}}]_1.
\end{aligned}
\end{equation}
This loss enables learning how to reconstruct similar images from the training data directly.

\noindent{\textbf{Adversarial Loss.}} The second term $\mathcal{L}_{adv}$ employs a discriminator $D$ for adversarial training~\cite{goodfellow2014generative}, which significantly improves the quality of the constructed image. Our model is to minimize it for $G$ and maximize it for $D$. Adversarial loss allows the generator to reconstruct the image as realistically as possible, while the discriminator distinguishes the normal images from the reconstructed and forgery abnormal images:
\begin{equation}
        \mathcal{L}_{adv} = \mathbb{E}_{\hat{\bm{I}} \sim p_{r}}[D({\hat{\bm{I}}})] + \mathbb{E}_{\widetilde{\bm{I}} \sim p_{f}}[D({\widetilde{\bm{I}}})] - \mathbb{E}_{\bm{I} \sim p_{n}}[D(\bm{I})].
\end{equation}

\noindent{\textbf{Latent Loss.}} Latent loss~\cite{akccay2019skip} penalizes the similarity between the positive and negative images in the latent space. In OCR-GAN, we use the features of the last convolutional layer of the discriminator $D$ as latent space features. Then, the $\ell_2$ error between latent space features is used as the latent loss:

\begin{equation}
    \mathcal{L}_{lat} = \mathbb{E}_{\bm{I}\sim p_n}[D_{lat}(\bm{I}) - D_{lat}(\hat{\bm{I}})]_2.
\end{equation}

Note that $\hat{\bm{I}} = G(\bm{I})$, $\widetilde{\bm{I}}$ is the forged abnormal data obtained by data augmentation, $p_{n}$, $p_{r}$ and $p_{f}$ are normal, reconstructed and forged image distributions, and $D_{lat}(\cdot)$ denotes the feature extraction of $D$ for the penultimate layer. The total loss $\mathcal{L}_{all}$ is a weighted sum of above losses:
\begin{equation}
    \mathcal{L}_{all} = \lambda_{con}\mathcal{L}_{con} + \lambda_{adv}\mathcal{L}_{adv} + \lambda_{lat}\mathcal{L}_{lat}.
\end{equation}
Based on the baseline skip-GANomaly settings and the results of our experiments, the weight parameters are chosen as $\lambda_{rec} = 50$, $\lambda_{adv} = 1$, and $\lambda_{lat} = 1$.

\subsection{Inference}
Anomaly score proposed in~\cite{schlegl2017unsupervised} is used to detect anomalies during inference.
For a test image $\bm{I}$, its anomaly score is defined as:
\begin{equation}
    A(\bm{I}) = \lambda \mathcal{L}_{con}(\bm{I})+(1-\lambda) \mathcal{L}_{lat}(\bm{I}),
\end{equation}
where $\mathcal{L}_{con}(\bm{I})$ is the reconstruction error that measures the content similarity between the input and reconstructed images, while $\mathcal{L}_{lat}(\bm{I})$ is the latent representation error based on the latent loss. Following the setting of our baseline skip-GANomaly, the weight parameter $\lambda$ is set to 0.9.

Based on Equation 13, we can compute the anomaly score for each test sample in the test set. The set of anomaly scores for all samples in the test set is $\bm{A}$. Following~\cite{akccay2019skip}, we scale $\bm{A}$ to $[0,1]$. Therefore, the final anomaly score for a test image $\bm{I}$ is:
\begin{equation}
    A^{'}(\bm{I}) = \frac{A(\bm{I}) - min(\bm{A})}{max(\bm{A}) - min(\bm{A})}.
\end{equation}

Threshold of anomaly score can be set according to requirements in real world applications.

\section{Experiments}\label{sec:experiments}
In order to assess the effectiveness of the proposed OCR-GAN, we consider two types of anomaly detection tasks: sensory AD and semantic AD. We evaluate the performance of the proposed OCR-GAN in two cases against state-of-the-art methods using public-available datasets. 
\subsection{Experimental Setup}
\subsubsection{Datasets}
This paper focuses on the sensory AD task. Thus, we use MVTec AD~\cite{bergmann2019mvtec}, DAGM~\cite{wieler2007weakly} and KolektorSDD~\cite{Tabernik2019JIM} to evaluate the performance of OCR-GAN in sensory AD. To further validate that OCR-GAN can improve the generation ability of generators in anomaly detection tasks, we use CIFAR-10~\cite{krizhevsky2009learning} to evaluate the performance of OCR-GAN in semantic AD.
 
\noindent\textbf{MVTec AD}~\cite{bergmann2019mvtec} contains 5,354 high-resolution color images that consist of 10 kinds of objects and 5 kinds of textures, which is widely used for the anomaly detection task. The image resolution ranges from 700 to 1,024, and we downscale all images to 256$\times$256 resolution for all experiments. The number of training samples for each category ranges from 60 to 320, and the abnormal samples in the test set contain more than 70 defects, \eg, cracks, scratches, deformation, and holes.

\noindent\textbf{DAGM}~\cite{wieler2007weakly} is a well-known benchmark database for surface defect detection. It contains images of various surfaces with artificially generated defects. Surfaces and defects are split into 10 classes of various difficulties. It is a weakly supervised dataset, and there are 8,050 training and testing sets each, and the ratio of positive and negative samples for each type is approximately 1:7. OCR-GAN is trained only on anomaly-free training samples for all experiments. 

\noindent\textbf{KolektorSDD}~\cite{krizhevsky2009learning} is constructed from images of defected electrical commutators. Specifically, microscopic fractions or cracks are observed on the surface of the plastic embedding in electrical commutators. The dataset contains 50 commutator samples, each with 8 surfaces, totaling 399 images in 500$\times$1,240, of which 347 images are without any defect, and 52 images are with visible defects. The anomalies are tiny and visually similar to the background, making this dataset challenging for anomaly detection.

\begin{table*}[htb]
	\centering
	\small
	\renewcommand\arraystretch{1.0}
	\renewcommand{\arraystretch}{1.1}
	\setlength\tabcolsep{5pt}
	\caption{\textbf{AUC results with SOTAs on MVTec AD dataset.} Three to ten columns are reconstruction-based methods while the following four columns are density-based and classification-based methods. \textbf{Bold} and \underline{underline} represent optimal and suboptimal results. ' means the year of publication. $^{\dag}$ means using the pre-trained model with extra dataset. $^{\ddag}$ means our training with forgery abnormal samples in Sec.~\ref{sec:as}.}
	\resizebox{1.0\textwidth}{!}{
	\begin{tabular}{C{5pt}|C{38pt}|C{23pt}C{20pt}C{21pt}C{24pt}C{24pt}C{24pt}C{23pt}C{25pt}|C{19pt}C{27pt}C{28pt}C{22pt}|C{27pt}C{28pt}}
		\hline
		&\multirow{3}{*}{Items} & {AGAN}& {AE$_1$}& {AE$_2$} & {SkipG} & {GradC} & {P-AE} & {DGAD} & {Draem} & {Diff} & {CutPaste} & {CutPaste$^\dag$} & {InTra} & \multirow{3}{*}{Ours} & \multirow{3}{*}{Ours$^\ddag$} \\
		~& & \cite{schlegl2017unsupervised} & \cite{bergmann2018improving} & \cite{bergmann2018improving} & \cite{akccay2019skip} & \cite{kwon2020backpropagated} & \cite{salehi2020puzzle} & \cite{xia2021discriminative} & ~\cite{zavrtanik2021draem} & \cite{rudolph2021same} & \cite{li2021cutpaste} & \cite{li2021cutpaste} & ~\cite{pirnay2022inpainting} & & \\ 
		~ & & 17' & 18' & 18' & 19' & 20' & 20' & 21' & 21' & 21' & 21' & 21' & 21' \\
		\hline
		\hline
		\multirow{6}{*}{\rotatebox{90}{texture}}&Carpet& 49.0 & 67.0 & 50.0 & 70.9 & 89.3 & 65.7 & 52.0 & 97.0 & 92.9 & 93.1 & \textbf{100.0} & 98.8 & 98.9\tiny{$\pm$0.5} & \underline{99.4}\tiny{$\pm$0.3} \\
		~&Grid& 51.0 &69.0 & 78.0 & 47.7 & 71.6 & 75.4 & 67.0 & \underline{99.9} & 84.0 & \underline{99.9} & 99.1 & \textbf{100.0} & 99.6\tiny{$\pm$0.2} & 99.6\tiny{$\pm$0.2} \\
		~&Leather& 52.0 & 46.0 & 44.0 & 60.9 & 69.3 & 72.9 & 94.0 & \textbf{100.0} & \underline{97.1} & \textbf{100.0} & \textbf{100.0} & \textbf{100.0} & \underline{97.1}\tiny{$\pm$0.6} & \underline{97.1}\tiny{$\pm$0.8} \\
		~&Tile& 51.0 & 52.0 & 77.0 & 29.9 & 63.4 & 65.5 & 83.0 & \underline{99.6} & 99.4 & 93.4 & \textbf{99.8} & 98.2 & 92.2\tiny{$\pm$0.8} & 95.5\tiny{$\pm$1.5} \\
		~&Wood& 68.0 & 83.0 & 74.0 & 19.9 & 76.7 & 89.5 & 72.0 & \underline{99.1} & \textbf{99.8} & 98.6 & \textbf{99.8} & 98.0 & 95.8\tiny{$\pm$1.6} & 95.7\tiny{$\pm$1.1} \\
		\hline
		~&\textbf{Average} & 54.2 & 63.4 & 64.6 & 45.86 & 74.1 & 73.8 & 73.6 & \underline{99.1} & 94.6 & 97.0 & \textbf{99.7} & 99.0 & 96.6\tiny{$\pm$0.3} & 97.5\tiny{$\pm$0.3} \\
		\hline
		\multirow{10}{*}{\rotatebox{90}{object}}&Bottle& 69.0 & 88.0 & 80.0 & 85.2 & 52.0 & 94.2 & 97.0 & 99.2 & 99.0 & 98.3 & \textbf{100.0} & \textbf{100.0} & \underline{99.6}\tiny{$\pm$0.2} & \underline{99.6}\tiny{$\pm$0.1} \\
		~&Cable& 53.0 &61.0 & 56.0 & 54.4 & 58.7 & 87.9 & 90.0 & 91.8 & 95.9 & 80.6 & 96.2 & 84.2 & \textbf{99.2}\tiny{$\pm$0.5} & \underline{99.1}\tiny{$\pm$0.6} \\
		~&Capsule& 58.0 & 61.0 & 62.0 & 54.3 & 55.6 & 66.9 & 60.0 & \textbf{98.5} & 86.9 & \underline{96.2} & 95.4 & 86.5 & 95.4\tiny{$\pm$0.4} & \underline{96.2}\tiny{$\pm$0.6} \\
		~&Hazelnut& 50.0 & 54.0 & 88.0 & 24.5 & 91.4 & 91.2 & 80.0 & \textbf{100.0} & 99.3 & 97.3 & \underline{99.9} & 95.7 & 88.2\tiny{$\pm$2.0} & 98.5\tiny{$\pm$1.3} \\
		~&Metal Nut& 50.0 & 54.0 & 73.0 & 81.4 & 56.0 & 66.3 & 95.0 & 98.7 & 96.1 & \underline{99.3} & 98.6 & 96.9 & 98.7\tiny{$\pm$0.2} & \textbf{99.5}\tiny{$\pm$0.3} \\
		~&Pill& 62.0 & 60.0 & 62.0 & 67.1 & 92.4 & 71.6 & 76.0 & \textbf{98.9} & 88.8 & 64.7 & 93.3 & 90.2 & \underline{98.5}\tiny{$\pm$0.4} & 98.3\tiny{$\pm$0.2} \\
		~&Screw& 35.0 & 51.0 & 69.0 & 87.9 & 78.2 & 57.8 & 67.0 & 93.9 & 96.3 & 86.3 & 86.6 & 95.7 & \textbf{100.0}\tiny{$\pm$0.0} & \textbf{100.0}\tiny{$\pm$0.0} \\
		~&Toothbrush& 57.0 & 74.0 & 98.0 & 58.6 & 98.0 & 97.8 & 93.0 & \textbf{100.0} & 98.6 & 98.3 & 90.7 & \textbf{99.7} & 98.2\tiny{$\pm$0.9} & 98.7\tiny{$\pm$0.7} \\
		~&Transistor& 67.0 & 52.0 & 71.0 & 84.5 & 72.8 & 86.0 & 88.0 & 93.1 & 91.1 & 95.5 & \underline{97.5} & 95.8 & 94.9\tiny{$\pm$0.3} & \textbf{98.3}\tiny{$\pm$1.5} \\
		~&Zipper& 59.0 & 80.0 & 80.0 & 76.1 & 56.6 & 75.7 & 82.0 & \textbf{100} & 95.1 & 99.4 & \underline{99.9} & 99.4 & 97.6\tiny{$\pm$0.4} & 99.0\tiny{$\pm$0.2} \\
		\hline
		~&\textbf{Average} & 56.0 & 63.5 & 73.9 & 67.4 & 71.2 & 79.5 & 82.8 & \underline{97.4} & 94.7 & 94.3 & 95.8 & 94.4 & 97.0\tiny{$\pm$0.2} & \textbf{98.7}\tiny{$\pm$0.3} \\
		\hline
		~&\textbf{All}& 55.0 & 63.0 &71.0 & 60.2 & 72.1 & 77.6 & 80.0 & \underline{98.0} & 94.7 & 95.2 & 97.1 & 95.9 & 96.9\tiny{$\pm$0.2} & \textbf{98.3}\tiny{$\pm$0.2} \\
		\hline
	\end{tabular}
	}
	\label{table:mvtec}
\end{table*}

\begin{table*}\normalsize
	\centering
	\small
	\renewcommand{\arraystretch}{1.1}
	\setlength\tabcolsep{8pt}
	\caption{\textbf{AUC results with SOTAs on DAGM dataset.} \textbf{Bold} and \underline{underline} represent optimal and suboptimal unsupervised results. *: original paper only reports average AUC, and the corresponding line shows our reproduced results.}
	\begin{tabular}{C{3pt} | c |C{20pt} C{20pt} C{20pt} C{20pt} C{20pt} C{20pt} C{20pt} C{20pt} C{20pt} C{23pt} |c}\hline
		\specialrule{0em}{1pt}{0pt}
		~ & Methods & Class1 & Class2 & Class3 & Class4 & Class5 & Class6 & Class7 & Class8 & Class9 & Class10 & Average\\
		\hline
		\multirow{6}{*}{\rotatebox{90}{Unsup.}} & skipGAN~\cite{akccay2019skip} & 58.3 & 56.1 & 55.1 & 53.7 & 57.4 & 66.8 & 52.4 & 53.7 & 52.3 & 52.2 & 55.8\\
		~ & Puzzle AE~\cite{salehi2020puzzle} & 50.7 & 50.5 & 58.7 & 70.0 & 63.6 & 92.3 & 54.0 & 49.1 & 54.6 & 49.6 & 59.3\\
		~ & CutPaste~\cite{li2021cutpaste} & 56.1 & 87.8 & 57.1 & 71.3 & 47.4 & 68.8 & 96.5 & 53.4 & 51.9 & 74.7 & 66.0\\
		~ & DifferNet~\cite{rudolph2021same} & 59.7 & 82.9 & 69.8 & 97.3 & 61.2 & 97.0 & 68.5 & 52.1 & 78.2 & 79.1 & 74.6\\
		~ & Draem~\cite{zavrtanik2021draem} & 96.1 & 98.3 & \textbf{99.5} & \textbf{99.6} & 92.1 & \textbf{100} & \underline{99.7} & \textbf{99.9} & \underline{98.9} & \underline{96.0} & \underline{98.0 ($99.0^*$)}\\
		%\hline
		% \rowcolor{mygray2}
		~ & \textbf{Ours} & \textbf{99.1} & \textbf{100} & \underline{99.1} & \underline{99} & \textbf{100} & \underline{97.5} & \textbf{99.8} & \underline{99.8} & \textbf{99.5} & \textbf{99.2} & \textbf{99.3}\\
		\hline
		\multirow{2}{*}{\rotatebox{90}{Sup.}} & Lin \textit{et al.}~\cite{lin2020efficient} & 100 & 94.0 & 100 & 100 & 100 & 100 & 100 & 99.0 & 100 & 100 & 99.3\\
		~ & B$\check{o}$ \textit{et al.}~\cite{bovzivc2021end} & 100 & 100 & 100 & 100 & 99.9 & 100 & 100 & 100 & 100 & 100 & 100\\
		\hline
	\end{tabular}
	\label{table:DGAM}
\end{table*}

\noindent\textbf{CIFAR-10}~\cite{krizhevsky2009learning} consists of 60,000 color images in 32$\times$32 with 10 classes. Semantic AD experiments regard one class as normal and the other classes as abnormal. CIFAR-10 is a challenging semantic AD dataset because images differ substantially across classes, and the background of images is not aligned. We experimented under two settings. Setting1 (S1) is protocol 2 described in \cite{perera2019ocgan}, which uses the whole training set of just one class as the normal data for training and the whole test set for the test time. Setting2 (S2) is the setting used in skip-GANomaly, that CIFAR-10 can yield 10 different anomaly cases, each with 45,000 normal training samples and 9,000:6,000 normal-anomaly test samples.

\subsubsection{Implementation Details.} Our method is implemented by PyTorch 1.2.0~\cite{paszke2019pytorch} and CUDA 10.2, and all experiments run with a TITAN RTX GPU. We use Adam~\cite{kingma2015adam} optimizer and set $\beta_1=0.5$, $\beta_2=0.999$, weight-decay=$1e^{-4}$, and learning rate=$0.002$. Unless otherwise specified, the batchsize is set to 32 for MVTec AD dataset, 64 for DAGM dataset, 64 for KolektorSDD dataset and 64 for CIFAR-10 dataset, and we use two frequencies (denoted as low-/high-frequency branches) for experiments. We choose Skip-GANomaly~\cite{akccay2019skip} as our baseline.

\begin{table}\normalsize
	\centering
	\small
	\renewcommand{\arraystretch}{1.1}
	\setlength\tabcolsep{8pt}
	\caption{\textbf{AUC results with SOTAs on KolektorSDD dataset.} \textbf{Bold} and \underline{underline} represent optimal and suboptimal results.}
	\begin{tabular}{c|c}\hline
		\specialrule{0em}{1pt}{0pt}
		Methods & AUC\\
		\hline
		skipGAN~\cite{akccay2019skip} & 55.1\\
		Puzzle AE~\cite{salehi2020puzzle} & 55.4\\
		DifferNet~\cite{rudolph2021same} & 84.9\\
		InTra~\cite{pirnay2022inpainting} & 70.1\\
		CutPaste~\cite{li2021cutpaste} & 60.2\\
		Draem~\cite{zavrtanik2021draem} & \underline{85.9}\\
		\hline
		Ours & \textbf{91.4}\\
		\hline
	\end{tabular}
	\label{table:KolektorSDD}
\end{table}

\begin{table*}\normalsize
	\centering
	\small
	\renewcommand{\arraystretch}{1.1}
	\setlength\tabcolsep{8pt}
	\caption{\textbf{AUC results with SOTAs on CIFAR-10 dataset.} \textbf{Bold} and \underline{underline} represent optimal and suboptimal results.}
	\begin{tabular}{c|c| c c c c c c c c c c |c}\hline
		\specialrule{0em}{1pt}{0pt}
		~&Methods & Plane & Car & Bird & Cat & Deer & Dog & Frog & Horse & Ship & Truck & Average\\
		\hline
		\multirow{10}{*}{\rotatebox{90}{S1}}&OCSVM~\cite{chen2001one} & 63.0& 44.0 & 64.9 & 48.7 & 73.5 & 50.0 & 72.5 &53.3 & 64.9 & 50.8 & 58.6\\
		~&AnoGAN~\cite{schlegl2017unsupervised} & 67.1 & 54.7 & 52.9 & 54.5 & 65.1 & 60.3 & 58.5 & 62.5 & 75.8 & 66.5 & 61.8\\
		% skipGAN~\cite{akccay2019skip} & 44.8 & \textbf{95.3} & 60.7 & 60.2 & 61.5 & \textbf{93.1} & \underline{78.8} & \textbf{79.7} & 65.9 & \textbf{90.7} & \underline{73.1}\\
        ~&skipGAN~\cite{akccay2019skip} & 65.6 & 47.6 & 66.0 & 57.8 & 74.6 & 58.7 & 61.6 & 64.7 & 76.1 & 69.1 & 64.2\\
		~&OCGAN~\cite{perera2019ocgan}  & 75.7 & 53.1 & 64.0 & \underline{62.0} & 72.3 & 62.0 & 72.3 & 57.5 & 82.0 & 55.4 & 65.7\\
		~&Gradcon~\cite{kwon2020backpropagated} & 76.0 & 59.8 & 64.8 & 58.6 & 73.3 & 60.3 & 68.4 & 56.7 & 78.4 & 67.8 & 66.4\\
		~&Puzzle AE~\cite{salehi2020puzzle} & \underline{78.9} & \underline{78.0} & \textbf{70.0} & 54.9 & \underline{75.5} & \underline{66.0} & \underline{74.8} & \underline{73.3} & \underline{83.3} & \underline{70.0} & \underline{72.5}\\
		~&Draem~\cite{zavrtanik2021draem} & 58.8 & 56.5 & 55.6 & 58.5 & 53.0 & 64.7 & 59.0 & 54.3 & 51.0 & 54.4 & 56.6\\
		%\hline
		~&CutPaste~\cite{li2021cutpaste} & 70.0 & 62.0 & 62.6 & \underline{62.0} & 53.8 & 62.9 & 62.4 & 59.9 & 51.8 & 57.6 & 60.5\\
		~&Intra~\cite{pirnay2022inpainting} & 50.2 & 48.9 & 57.8 & 49.2 & 55.4 & 60.3 & 44.5 & 65.7 & 73.8 & 64.9 & 57.1\\
		%\hline
		% Ours & \textbf{99.9} & \underline{80.2} & \textbf{82.5} & \textbf{85.4} & \textbf{98.5} & \underline{87.3} & \textbf{98.6} & \underline{76.9} & \textbf{99.8} & \underline{85.2} & \textbf{89.4}\\
          ~&\textbf{Ours} & \textbf{82.0} & \textbf{78.9} & \underline{68.4} & \textbf{0,0,1}{78.9} & \textbf{84.5} & \textbf{80.8} & \textbf{75.1} & \textbf{89.6} & \textbf{84.4} & \textbf{72.2} & \textbf{79.5}\\
		\hline
    \multirow{2}{*}{\rotatebox{90}{S2}} &skipGAN~\cite{akccay2019skip} & 44.8 & \textbf{95.3} & 60.7 & 60.2 & 61.5 & \textbf{93.1} & \underline{78.8} & \textbf{79.7} & 65.9 & \textbf{90.7} & \underline{73.1}\\
    ~&\textbf{Ours} & \textbf{99.9} & \underline{80.2} & \textbf{82.5} & \textbf{85.4} & \textbf{98.5} & \underline{87.3} & \textbf{98.6} & \underline{76.9} & \textbf{99.8} & \underline{85.2} & \textbf{89.4}\\
    \hline
	\end{tabular}
	\label{table:cifar10}
\end{table*}

\subsubsection{Evaluation Metrics}
The Area Under the Curve (AUC) of the Receiver Operating Characteristic (ROC) Curve is used as a standard evaluation metric for anomaly detection. It is calculated by gradually changing the threshold of anomaly scores. AUC is accumulated to a score for the performance evaluation. A higher AUC score means better anomaly detection performance.

\subsubsection{Default Setting}
Following the previous unsupervised anomaly detection methods settings, our ablation experiments and interpretability experiments are all conducted on the MVTec AD dataset unless otherwise specified. Considering the number of parameters and the computational cost, we choose two frequency branches (high frequency and low frequency) for the experiments if not specified. Moreover, our OCR-GAN keeps the same parameter settings as the baseline. If not specified, the number of feature channels of each generator is set to 64.

\subsection{Compare with SOTA Methods}
We evaluate the performance of our OCR-GAN on four popular datasets to verify the superiority of our method over other SOTA methods.

\noindent\textbf{1) Sensory AD.} The difference between normal and abnormal images in the sensory AD task is covariate shift. And anomalies usually appear in the form of defects, such as cracks, scratches and holes. Actually, we design our OCR-GAN based on the motivation of difference in the frequency distribution of normal and abnormal samples in the MVTec AD dataset (shown in Fig.~\ref{fig:motivation}(a)). And the frequency analysis can be extended to other sensory AD datasets. Therefore, we choose three different sensory AD datasets to evaluate the effectiveness of our method in the sensory AD task.

\noindent\textbf{MVTec AD.} Tab.~\ref{table:mvtec} shows the detection AUC results of different methods on the MVTec AD dataset, and our experiments run five times using different random seeds without extra training data. We report the mean AUC score with corresponding standard error for each category and the average AUC for texture, object, and all categories. Results indicate that our approach achieves a new SOTA on the MVTec AD dataset without extra training data, \ie, obtaining 98.3 detection AUC score.
OCR-GAN improves the AUC score by a significant +18.3$\uparrow$ compared with SOTA classical reconstruction-based method DGAD~\cite{xia2021discriminative} without extra training data, by 3.6$\uparrow$ compared with SOTA density-based method DifferNet~\cite{rudolph2021same}, and by 1.2$\uparrow$ compared with SOTA classification-based method CutPaste~\cite{li2021cutpaste}. Our OCR-GAN belongs to classical reconstruction-based approaches that use the generator to reconstruct the image and use the reconstruction error to detect anomalies. The current SOTA method, Draem~\cite{zavrtanik2021draem}, trains two models (reconstruction model and anomaly segmentation model) using extra training data and uses the results of the anomaly segmentation to detect anomalies, which is very different from the classical reconstruction-based approaches. However, our OCR-GAN achieves a higher AUC score than Draem~\cite{zavrtanik2021draem} by +0.3 $\uparrow$ without using extra training data, proving the effectiveness of our approach. Although our OCR-GAN does not achieve the best performance in every category, we obtain the highest overall score, and the AUC score of each category surpasses 95 (\cf Ours$^{\ddag}$ in the table) proving the robustness and practicality of our method. In conclusion, it is remarkable that we first achieve SOTA on the MVTec AD dataset with a classical reconstruction-based method without extra training data while previous classical reconstruction-based methods fail to perform so well on sensory AD.

\noindent\textbf{DAGM.} Our OCR-GAN is trained only on normal samples for the DAGM dataset.
Experimental results of other unsupervised methods on DAGM are obtained by experimenting using their open-source code. Tab.~\ref{table:DGAM} shows that supervised methods have achieved near-perfect AUC on the DAGM dataset. Compared with supervised methods, previous unsupervised methods (including classification-based methods, reconstruction-based methods, and density-based methods) did not perform well on this dataset. However, our OCR-GAN achieves a 99.3 detection AUC score without extra training data. Our performance is comparable to supervised methods, which is a remarkable result.

\noindent\textbf{KolektorSDD.} OCR-GAN is compared with other unsupervised methods, and results are shown in Tab.~\ref{table:KolektorSDD}. As the anomaly elements in the dataset are small and similar to the background, previous unsupervised methods do not perform well on the KolektorSDD dataset. Without extra training data, our OCR-GAN achieves a 91.4 detection AUC score that increases by +5.5$\uparrow$ over the suboptimal method Draem which is trained with extra data. This result shows that our approach also performs reliably in challenging sensory AD datasets.

\noindent\textbf{2) Semantic AD.} Experiments on the semantic anomaly detection task are performed to assist in proving the effectiveness of our approach. The difference between normal and abnormal images in the semantic AD~\cite{perera2019learning} task is label shift. There is no variability across frequency bands of normal and abnormal in this task. However, our method can also improve the generation ability of the generator. We choose a public dataset CIFAR-10, which is widely used for one-class detection (semantic AD), to verify the effectiveness of our method on all types of anomaly detection tasks.

\noindent\textbf{CIFAR-10.} As shown in Tab.~\ref{table:cifar10}, reconstruction-based methods perform better in the semantic AD compared to the density-based methods and the classification-based methods. 1) In the experiment under Setting1, our OCR-GAN outperforms all compared unsupervised methods and achieves 79.5 AUC on the CIFAR-10 dataset, which is +7.0$\uparrow$ higher than the suboptimal method. 2) In the experiment under Setting2, the performance of our method is greatly improved +16.3$\uparrow$ compared to the baseline skip-GANanomaly. The results verify that our OCR-GAN performs well in one-class detection (semantic AD). Although the difference in frequency bands is not suitable for semantic anomaly detection, the improvement to structural design implicitly improves the model's reconstruction ability. Therefore the performance of the model in semantic AD is also enhanced.

\subsection{Ablation Study} \label{sec:as}
\begin{table}[t]\normalsize
	\centering
	%\small
	\renewcommand{\arraystretch}{1.1}
	\setlength\tabcolsep{7pt}
	\caption{Ablation study on MVTec AD dataset. BN represents branch number.}
    \begin{tabular}{c c c c c c |l}
    \hline
    \specialrule{0em}{1pt}{1pt}
    No.&\makecell[c]{BN} & FD & CS & cutout & cutpaste & AUC\\ 
    \hline
    (1)&\color{mygray3}2&{\color{mygray3}\ding{55}}&{\color{mygray3}\ding{55}}&{\color{mygray3}\ding{55}}&{\color{mygray3}\ding{55}}&60.2\\
    (2)&\color{mygray3}2&{\color{mygray3}\ding{55}}&{\color{mygray3}\ding{55}}&\ding{51}&\ding{51}&68.5\tiny{+8.3}\\
    (3)&\color{mygray3}2&\ding{51}&{\color{mygray3}\ding{55}}&{\color{mygray3}\ding{55}}&{\color{mygray3}\ding{55}}&74.8\tiny{+14.6}\\
    (4)&\color{mygray3}2&\ding{51}&{\color{mygray3}\ding{55}}&\ding{51}&\ding{51}&77.2\tiny{+17.0}\\
    (5)&\color{mygray3}2&\ding{51}&\ding{51}&{\color{mygray3}\ding{55}}&{\color{mygray3}\ding{55}}&96.9\tiny{+36.7}\\
    (6)&3&\ding{51}&\ding{51}&{\color{mygray3}\ding{55}}&{\color{mygray3}\ding{55}}&97.5\tiny{+37.3}\\
    (7)&\color{mygray3}2&\ding{51}&\ding{51}&\ding{51}&{\color{mygray3}\ding{55}}&97.4\tiny{+37.2}\\
    (8)&\color{mygray3}2&\ding{51}&\ding{51}&{\color{mygray3}\ding{55}}&\ding{51}&97.5\tiny{+37.3}\\
    (9)&\color{mygray3}2&\ding{51}&\ding{51}&\ding{51}&\ding{51}&98.3\textcolor{red}{\tiny{+38.1}}\\
    \hline
    \end{tabular}
    \label{table:Ablation}
\end{table}
\noindent\textbf{Influence of Different Components.} We further conduct an ablation study on the MVTec AD dataset to investigate the effectiveness of each component of the proposed OCR-GAN. We choose Skip-GANomaly~\cite{akccay2019skip} as our baseline and gradually add the different component, performing the following seven experiments: \textbf{\emph{(1)}} Baseline; \textbf{\emph{(2)}} Baseline training with forgery abnormal images by both cutout and cutpaste data augmentations; \textbf{\emph{(3)}} Adding FD module; \textbf{\emph{(4)}} (3) training with forgery abnormal images by both cutout and cutpaste data augmentations; \textbf{\emph{(5)}} Adding both FD and CS modules, \ie, OCR-GAN in the paper; \textbf{\emph{(6)}} Using three frequency branches; \textbf{\emph{(7)}} OCR-GAN training with forgery abnormal images by the cutout data augmentation; \textbf{\emph{(8)}} OCR-GAN training with forgery abnormal images by the cutpaste data augmentation; \textbf{\emph{(9)}} OCR-GAN training with forgery abnormal images by both cutout and cutpaste data augmentations. As shown in Tab.~\ref{table:Ablation}, our baseline only obtains a 60.2 AUC score because this reconstruction-based method suffers from the poor reconstruction ability of the generator. Training with forgery abnormal images can bring +8.3$\uparrow$ AUC improvement to the baseline model. When the FD module is added to the baseline, the model performance increases by a significant +14.6$\uparrow$, and our proposed CS module further improves the AUC score by +22.1$\uparrow$ to 96.9. And adding these two modules to the baseline with cutout/cutpaste still gives a significant improvement in anomaly detection performance, which verifies that the main improvement of the model is from FD and CS. The results strongly demonstrate the effectiveness of our proposed two modules for the anomaly detection task. Moreover, our approach improves by +0.6$\uparrow$ when using three frequency branches, meaning that more frequencies contribute to the model performance. We set the frequency number as two in the paper to balance model effectiveness and efficiency. And we use data augmentation to generate forgery abnormal samples to assist the training process. As shown in Tab.~\ref{table:Ablation}, each data augmentation contributes to the model performance, and our OCR-GAN obtains the best result when both augmentations are applied.
\begin{figure*}[t]
    \centering
    \includegraphics[width=0.8\textwidth]{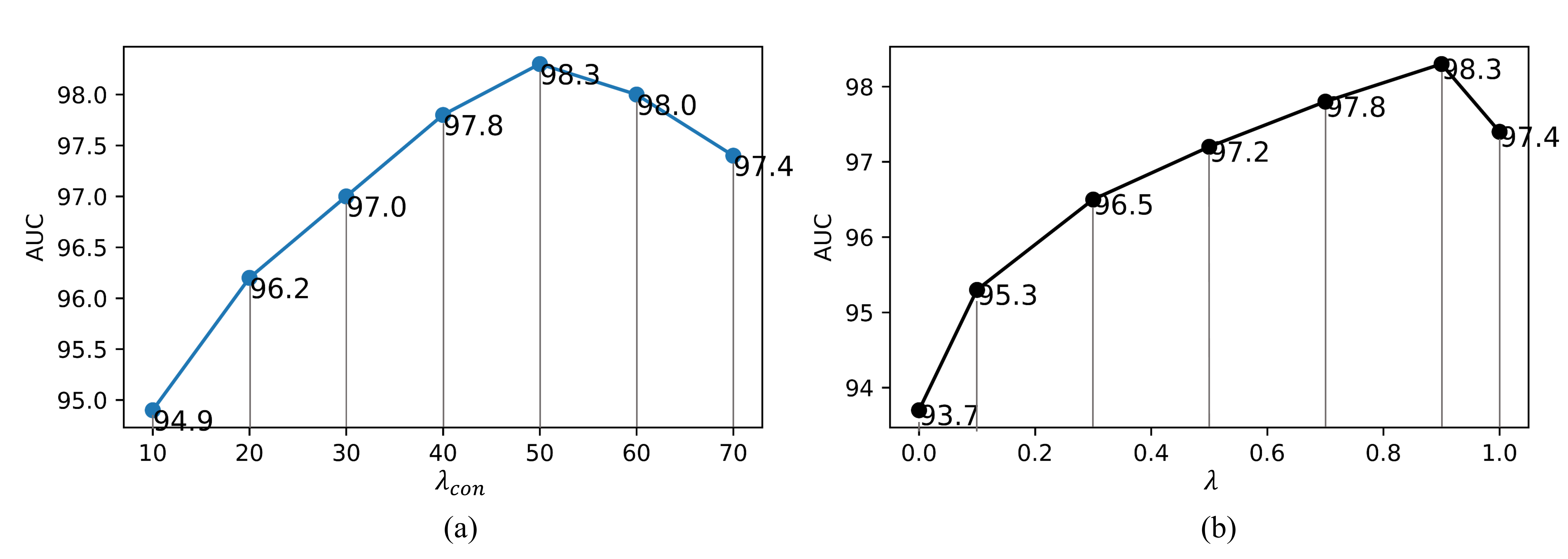}
    \caption{(a): Ablation study for the value of $\lambda_{con}$ in loss function; (b): Ablation study for the value of lambda in anomaly score.}
    \label{fig:lambda}
\end{figure*}

\begin{table}[t]\normalsize
    \small
    \centering
	\renewcommand{\arraystretch}{1.17}
    \setlength\tabcolsep{6pt}
    \caption{\centering Ablation study for frequency branches.}
    \begin{tabular}{c |c c c c}
    \hline
    \specialrule{0em}{1pt}{1pt}
    Category &\makecell[c]{high\\frequency} & \makecell[c]{low\\frequency} & \makecell[c]{two\\branches} & OCR-GAN\\ 
    \hline
    texture& 85.2 & 69.8 & 73.6 & 96.6\\
    object& 81.3 & 75.1 & 75.4 & 97.0 \\
    \hline
    all& 82.6 & 73.3 & 74.8 & 96.9\\
    \hline
    \end{tabular}
    \label{table:Ablation_band}
\end{table}

\noindent\textbf{Influence of Frequency Branches.} Flat distribution corresponds to the low-frequency component, while sharp changes (e.g., edge and noise) denote the high-frequency term. As different frequencies contain different information, we conduct an ablation study on the MVTec AD dataset to explore the anomaly detection performance using different frequency branches. As shown in Tab.~\ref{table:Ablation_band}, we conduct a set of experiments using only a high-frequency branch, only a low-frequency branch, two independent frequency branches, and two frequency branches with CS module (two-frequency-branch OCR-GAN). Results show that using only high-frequency information performs better than low-frequency information, meaning that abnormal elements contain more high-frequency information. Nevertheless, using two-frequency branches independently is not ideal for lacking the information interaction between different frequency branches. Our designed CS module can well handle this problem and further improve the model performance.
\begin{table}[!t]\normalsize
\centering
\caption{Ablation study of using different loss components.}
\begin{tabular}{ccc|c}
\hline
$\mathcal{L}_{con}$ & $\mathcal{L}_{adv}$ & $\mathcal{L}_{lat}$ & AUC \\ \hline
\ding{51}&{\color{mygray3}\ding{55}}&{\color{mygray3}\ding{55}}& 96.3\\ 
{\color{mygray3}\ding{55}}&\ding{51}&{\color{mygray3}\ding{55}}& 90.5\\
{\color{mygray3}\ding{55}}&{\color{mygray3}\ding{55}}&   \ding{51}& 94.5\\ 
\ding{51}&\ding{51}&{\color{mygray3}\ding{55}}& 97.1\\ 
{\color{mygray3}\ding{55}}&\ding{51}&\ding{51}& 95.0\\ 
\ding{51}&{\color{mygray3}\ding{55}}&\ding{51}& 97.6\\ 
 \ding{51}&\ding{51}&\ding{51}&98.3\\ \hline
% \begin{tabular}{c|cccccccc}
% \hline
% $\lambda_{con}$   & 1 & 0 & 0 & 1 & 10 & 40 & 50 & 60 \\ \hline
% $\lambda_{adv}$   & 0 & 1 & 0 & 1 & 1  & 1  & 1 & 1 \\ \hline
% $\lambda_{lat}$   & 0 & 0 & 1 & 1 & 1  & 1  & 1 & 1 \\ \hline
% AUC & 97.0  &  90.5 & 94.5  &  94.2 &  95.4  &  97.7  & 98.3 & 97.2  \\ \hline
\end{tabular}
\label{table:loss}
\end{table}

\noindent\textbf{Influence of hyperparameter values.} 
The training and testing process of our OCR-GAN involves the choosing of several hyperparameters, such as weighting parameters for different losses in Equation(12) and weighting parameters for reconstructed differences and latent space differences in Equation(13). The choice of these parameters in our experiments follows the parameter settings of our baseline skip-GANomaly. We further conduct experiments to explore the influence of the value of the hyperparameters on the anomaly detection performance of the model.
We conduct the ablation experiments about the parameter settings in Equation(11). Fig.~\ref{fig:lambda}(a) shows how the value of $\lambda_{con}$ in the loss function influences the model performance, and Tab.~\ref{table:loss} shows the ablation study of using different loss components. The results show the contribution of each loss component to the anomaly detection performance.
We also conduct experiments to explore the influence of the value of lambda in Equation(13). As shown in Fig.~\ref{fig:lambda}(b), when this parameter is set to 0.9, the model achieves the best anomaly detection performance. 

% \begin{figure}[t]
%     \centering
%     \includegraphics[width=1.0\columnwidth]{figs/lambda.pdf}
%     \caption{\textcolor[rgb]{0,0,1}{Ablation study for the value of lambda in anomaly score.}}
%     \label{fig:lambda}
% \end{figure}

\begin{figure*}[t]
    \centering
    \includegraphics[width=1\textwidth]{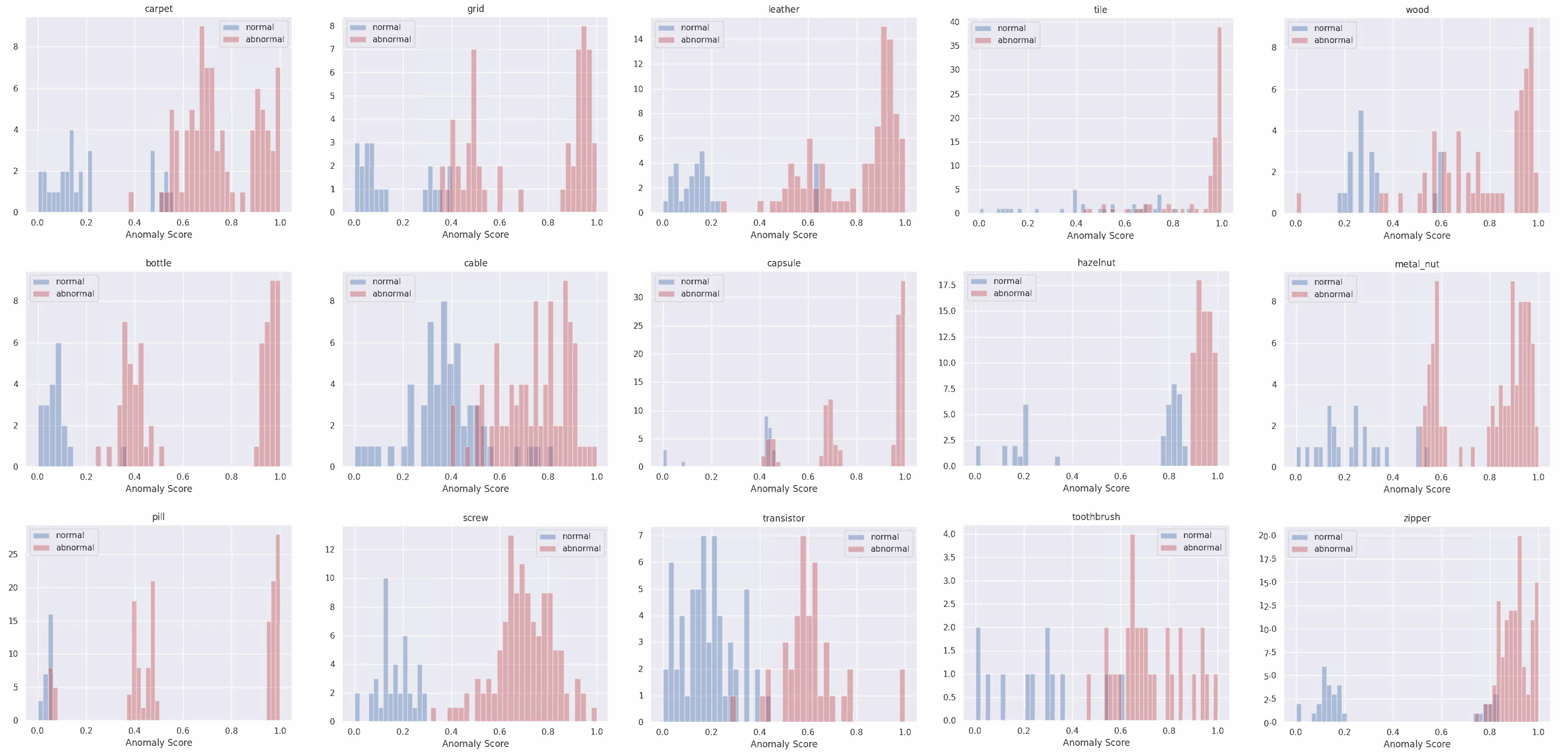} 
    \caption{Histogram of anomaly scores for the normal and abnormal samples for each category in the MVTec AD dataset.}
    \label{fig:scores}
\end{figure*}
\begin{figure*}[t]
	\centering
	\includegraphics[width=1\textwidth]{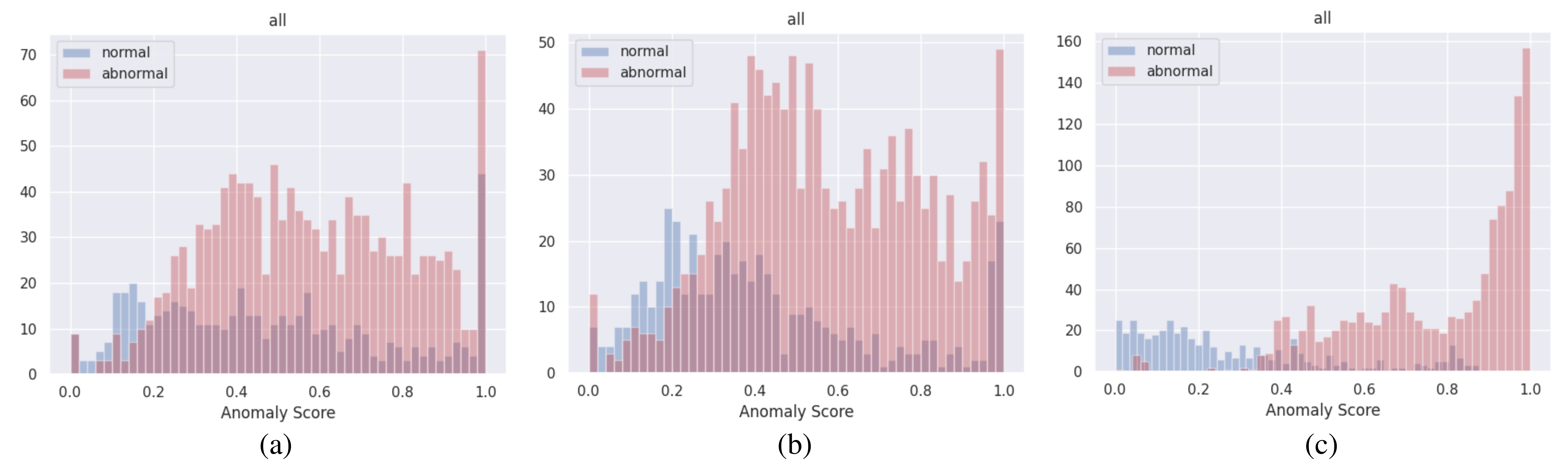}
	\caption{Comparison of anomaly score histograms for all category. \textbf{(a)}:Baseline. \textbf{(b)}:Adding FD. \textbf{(c)}:Adding both FD and CS.}
	\label{fig:com_score}
\end{figure*}

\begin{figure*}[tp]
	\centering
	\includegraphics[width=1\textwidth]{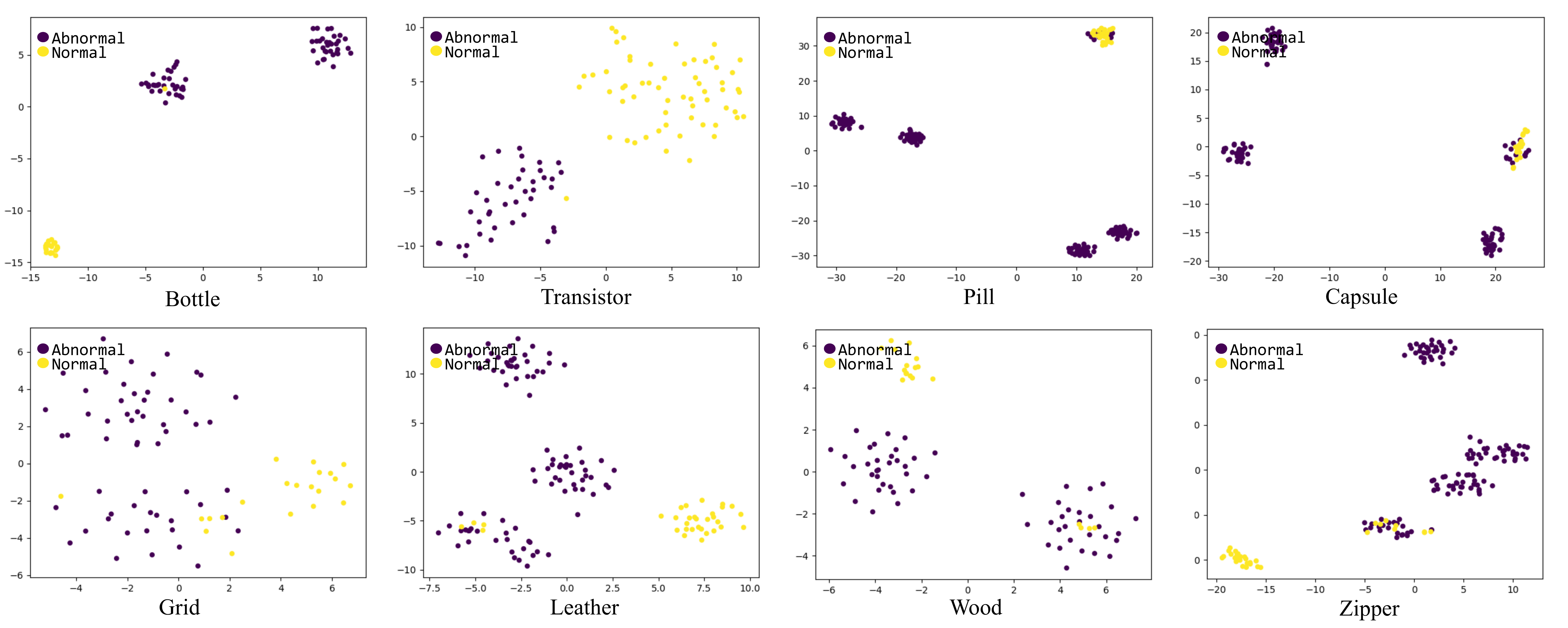}
	\caption{\textbf{t-SNE visualization} of normal and abnormal samples for eight categories in MVTec AD dataset.}
	\label{fig:tsne}
\end{figure*}

\begin{figure}[t]
	\centering
	\includegraphics[width=1\columnwidth]{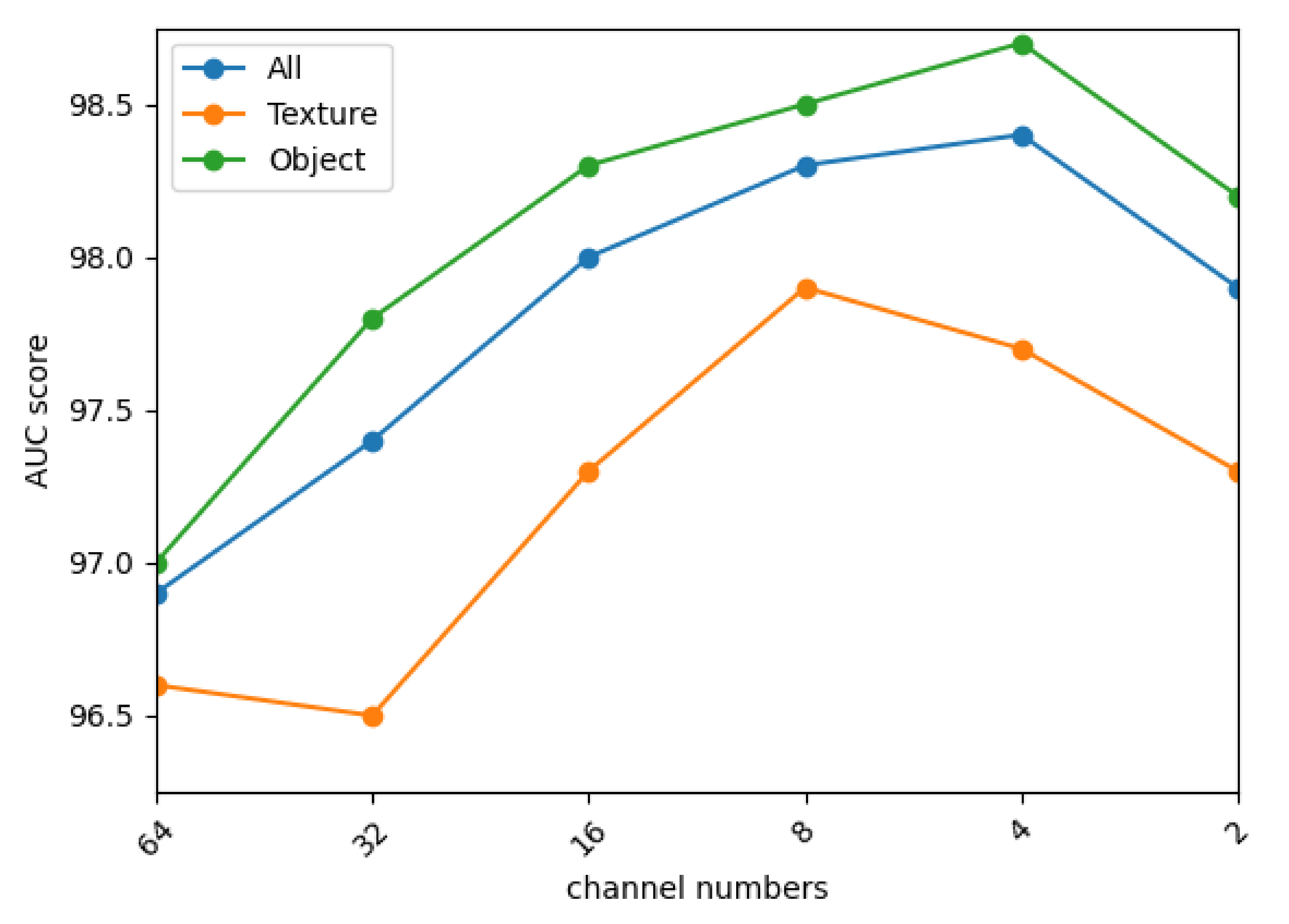}
	\caption{\textbf{Influence of channel numbers.} With the number of generator feature channels decreases, the AUC first increases and then decreases.}
	\label{fig:channel}
\end{figure}

\noindent\textbf{Influence of Number of Parameters.} Our OCR-GAN reconstructs the information of different frequency bands respectively, which means that one frequency band requires one generator. Considering the model efficiency, we conduct experiments mostly with two frequency branches. Compared with the baseline, our OCR-GAN has a larger number of method parameters. So, we reduce the number of feature channels of the generator to explore the influence of model parameters on the model performance. As shown in Fig.~\ref{fig:channel}, when we reduce the number of feature channels, the model performs better. \emph{Our model performs best when the number of channels is set to 4, achieving 98.7 AUC on the MVTec AD dataset. This means that our lightweight model can even perform better in anomaly detection task.} Since the lightweight model is not the focus of this study, we will further fully study the design of the lightweight anomaly detection model and why the lightweight model can bring certain performance improvements in future work. For a fair comparison, all of our experiments in this paper (except this one) use the standard model, \ie, the number of feature channels equals 64. 

\subsection{Interpretability of OCR-GAN}
\noindent\textbf{Analysis of Histogram.}
We visualize the anomaly score histogram for each category to further prove the effectiveness of our OCR-GAN. As shown in Fig.~\ref{fig:scores}, abnormal samples would get higher anomaly scores while normal samples get lower anomaly scores, and there is a clear distinction between normal and abnormal samples, meaning that our model can well distinguish abnormal samples from normal samples by the anomaly score. Fig.~\ref{fig:com_score} shows that normal samples and abnormal samples can not be distinguished by anomaly score in the histogram of the baseline. We further assess our proposed FD and CS module and results similarly indicate that each module contributes to the model.

\noindent\textbf{Visualization of latent-space features. }
We map the latent-space features from the last convolution layer of the $D$ for each test sample to a two-dimensional subspace. Fig.~\ref{fig:tsne} shows that our proposed OCR-GAN yields promising separation between normal and abnormal samples in the latent space.

\begin{figure}[tp]
	\centering
	\includegraphics[width=1\columnwidth]{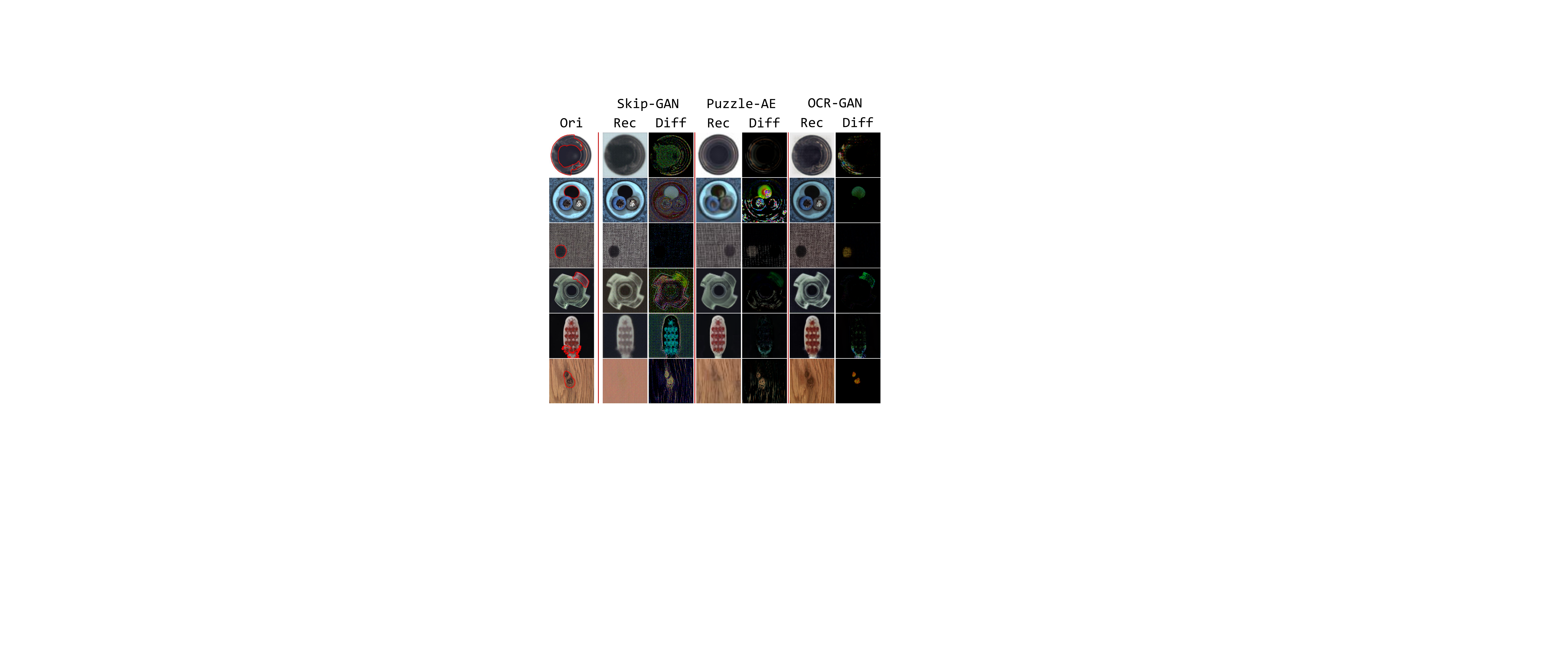}
	\caption{\textbf{Reconstruction results of three reconstruction-based methods.} Ori: Original images with anomaly segmentation ground truth. Rec: Reconstructed images. Diff: Differences between original and reconstructed images.}
	\vspace{-0.5em}
	\label{fig:recon}
\end{figure}

\noindent\textbf{Reconstruction results. }
The reconstruction ability of the generator has a significant influence on the performance of the GAN-based method in the anomaly detection task. As shown in Fig.~\ref{fig:recon}, we visualize the reconstructed images and the difference images between the reconstructed and original images to explore the reconstruction ability of different methods. Reconstructed results indicate that our OCR-GAN has a better reconstruction ability for details, and the abnormal areas are more prominent in the difference images than other classical reconstruction-based methods.

\section{Conclusion}\label{sec:conclusion}
This paper proposes a novel reconstruction-based OCR-GAN for anomaly detection from the perspective of the frequency domain. Specifically, we propose FD module to decouple the input image into different frequencies and model the reconstruction process as a combination of parallel omni-frequency image restorations. To better perform frequency interaction among different encoders, we propose a tailored CS module to adaptively select different channels among multiple branches. Our approach achieves new SOTA results over current SOTA methods on both sensory AD and semantic AD tasks even without extra training data, meaning that the proposed OCR-GAN is robust and effective for practical applications. 

In the future, we will further explore the design of the lightweight model for AD tasks, while building a more difficult practical dataset and testing the corresponding effects of our and other methods, hoping to contribute to the development of this field.

\section*{Acknowledgements}\label{sec:acknowledgements}
We thank all reviewers and the editor for excellent contributions. This work is supported by the the National Natural Science Foundation of China (NSFC) under the Grant No. U21A20484 and the Key Research and Development Program of Zhejiang Province (No. 2022C01011).

\bibliographystyle{IEEEtran}
\bibliography{main}

\end{document}